\documentclass[runningheads]{llncs}
\usepackage[T1]{fontenc}
\usepackage{graphicx}
\usepackage{natbib}
\usepackage{multirow}
\usepackage{color}
\usepackage{support-caption}
\usepackage{subfig}
\usepackage{float}
\usepackage[colorlinks,bookmarks=false,linkcolor=blue,urlcolor=blue, citecolor=blue]{hyperref}
\begin{document}
\title{Needle In A Haystack, Fast: Benchmarking Image Perceptual Similarity Metrics At Scale}
%
%
\author{Cyril Vallez\inst{1,2} \and
 Andrei Kucharavy\inst{1}\orcidID{0000-0003-0429-8644} \and
Ljiljana Dolamic\inst{2}\orcidID{0000-0002-0656-5315}}
\authorrunning{Vallez et al.}
\titlerunning{Image Perceptual Similarity Metrics at Scale}
%
\institute{Ecole Polytechnique Federale Lausanne, Route Cantonale 1015, Lausanne, Switzerland\\
\email{\{firstname.name\}@epfl.ch}\\
\and
Cyber-Defence Campus, armasuisse, Feuerwerkerstrasse 39, 3602 Thun, Switzerland\\
\email{\{firstname.name\}@armasuisse.ch}\\
}
\maketitle              
\begin{abstract}
The advent of the internet, followed shortly by the social media made it ubiquitous in consuming and sharing information between anyone with access to it. The evolution in the consumption of media driven by this change, led to the emergence of images as means to express oneself, convey information and convince others efficiently.
With computer vision algorithms progressing radically over the last decade, it is become easier and easier to study at scale the role of images in the flow of information online. While the research questions and overall pipelines differ radically, almost all start with a crucial first step - evaluation of global perceptual similarity between different images.
That initial step is crucial for overall pipeline performance and processes most images. A number of algorithms are available and currently used to perform it, but so far no comprehensive review was available to guide the choice of researchers as to the choice of an algorithm best suited to their question, assumptions and computational resources.
With this paper we aim to fill this gap, showing that classical computer vision methods are not necessarily the best approach, whereas a pair of relatively little used methods - Dhash perceptual hash and SimCLR v2 ResNets achieve excellent performance, scale well and are computationally efficient.

\keywords{Image fingerprinting \and Perceptual Similarity \and Deep Learning \and Image similarity Metrics \and Benchmarking \and Computer vision.}
\end{abstract}

\section{Introduction}

\subsection{Image role in the information flow in (social) media}

The arrival of the wide-spread internet availability, followed by the arrival of the social media shortly after has made them ubiquitous in the information sharing and consumption by anyone with access to internet.

Arguably thrusted into the attention of general scientific communities by the role they played in the 2011 Arab Spring \citep{ArabSpringTwitter2014}, the analysis of information flowing through social media have since been extensively explored, ranging from understanding aboriginal communities \citep{AboriginalCommunities2022}, to self-depreciative humor of on-campus students reflective of their mental health issues and more general struggles \citep{StudentDepreciativeHumor2018}. 

Building on top of the original catalyst, a substantial amount of research has focused on investigating information flows related to the political issues, ranging from American Indigenous groups social media activism in the context of oil pipeline construction  \citep{NoDAPLTwitter2019} to long-term flops and successes around the communication about the anthropogenic climate change. \citep{SocialMediaClimateChange2019}.

While initially the research focused on the textual content of the information shared, extensive evidence has emerged that visual information, such as pictures, memes, diagrams or infographics were just as if not more important for the information flow on social media \citep{Instagrammatics2016, VisualMethodsPearce2020}. In fact, their importance is so high that they have been repeatedly used as vectors of disinformation and misinformation, first in traditional media (such as 2004 election photo edit to present John Kerry as anti-war activist - \citep{lightfonda2004}), and then on the social media, with notable first incident being Hurricane Sandy \citep{SandyHurricaneFakes2013} and one of the latest development being the 2020 US presidential election fraud allegations \citep{abilov2021voterfraud2020}. 

In response to such a proliferation of images, numerous methods to detect malicious image manipulations, re-use or misattributions have been proposed. Some of them are 
DejaVu collaborative deceptive image index developed by \citet{dejavu2018matatov},
Image Phylogeny by \citet{ProvenancePhilogeny2017PintoScheirer},
Image Provenance Analysis by \citet{Provenance2018MoreiraScheirer},
Hybrid Image Retrieval by \citet{HybridImageRetrieval2019PhamPark},
MemeSequencer for parsing of memes by \citet{memesequencer2020dubey}, 
Transformation-Aware image embedding \citet{TransformAwareImageEmbedding2021BhartiScheirer}, 
and lately 
Motif Mining for detection of remixed content, such as photoshop insertions, by \citet{MotifMining2022} 
and Visual Almost Duplication Detection by \citet{NearIdenticalVisualImages2022Matatov}.

While enabled at scale by efficient feature search and matching, made possible notably by \citet{FAISS2019Facebook}, they all share one critical first step - namely global perceptual image feature extraction. While the pipelines presented above might be specialized, the task is more general and a variety of radically different methods are currently employed to transform images to perceptual features.

To our knowledge, however, there has not been any reviews how different families of approaches to such perceptual feature extraction compare.

This is highly problematic, given that most research questions regarding information flow on social media originate outside the computer vision field. Such research requires substantial domain-specific expertise to properly formulate and establish rules for data collection \citep{HashtagsMatter2020Geboers}, meaning few means are available to investigate sate-of-the-art algorithms in computer vision.

Hence this review focuses on visual similarity search methods that are commonly used and readily available as python libraries.

\subsection{Perceptual similarity features extraction methods}

In itself, the annotation of the content of image and search for perceptually or semantically similar images is one of the core problems in the computer vision. Early attempts to solve those issues are well reviewed by \citet{1973DudaHart}. 




\subsubsection{Local keypoint matching algorithms}

A core problem with image comparison is that features detected by a human eye poorly translate into the pixel array representation of images stored by computers and vice-versa. One of the first widely successful approaches was to find unique groups of pixels that were unique and highly distinctive between images (eg. corners of eyes, mouth and nose for face detection) and use sets of them (bags-of-features) to describe and compare images. Algorithms based on such local feature detectors are known as local keypoint matching.

Despite having been initially described and shown to be effective in the late 80s and early 90s, they are still widely used in state-of-the art image similarity detection. The \textit{Scale-Invariant Feature Transform} SIFT \citep{SIFT} is arguably the most known representative of this group and early on has been shown to be one of the best performing keypoint detectors out there \citep{SIFT_FTW2005}. Based on sampling a pattern of luminosity around a point it had one main default - a relatively low speed. To accelerate it, several approaches were introduced, such as \textit{Speed-Up Robust Features} SURF \citep{SURF2008}. Both of them are used respectively by the Hybrid Image Retrieval pipeline \citep{HybridImageRetrieval2019PhamPark} and by Image Phylogeny/Image Provenance Analysis pipelines \citep{ProvenancePhilogeny2017PintoScheirer, Provenance2018MoreiraScheirer}.

Another notable local feature is \textit{FAST} corner detection algorithm \cite{FAST1998}. Unlike SIFT and its derivatives, rather than looking at pixel patterns, it uses pixel intensity over a circular pattern to detect corners. Significantly faster than SIFT, it is at the base of \textit{DAISY} and \textit{LATCH} methods. DAISY that uses a log-polar sampling of FAST keypoints to create unique features \citep{DAISY}, whereas LATCH uses relative position of triplets of keypoints extracted by FAST to create unique features \citep{LATCH}. A slightly different approach is used by \textit{Oriented FAST and Rotated BRIEF} ORB \citep{ORB}, where the detection of keypoints is still done by FAST, but features are built by selecting a set of random pairs of points detected by FAST and comparing them, resulting in a binary string. Each set of random points is a feature and the approach itself is known as BRIEF descriptor \citep{BRIEF2010}, plus a modification to make sure the descriptor is robust to rotations. All of them are in use in modern image similarity search, with ORB, combined with the principal component analysis (PCA), being notably used by DejaVu \citep{dejavu2018matatov}.

Numerous other methods for keypoint extraction, feature description and feature matching between images have been developed. An excellent review on the subject is available from \citet{FeatureExtractionSurvey2021}.

A critical feature of all the keypoint extraction and comparison methods is that they allow to exactly identify identical regions between two images, notably in case of a major viewpoint change. However, both the exact region-region match and viewpoint change detection are necessarily the deformations we expect to find in media shared and re-shared on social medias.




\subsubsection{Block perceptual hashing}

In order to achieve the ability to match regions and a robustness with regards to a viewpoint change, keypoint-based detectors tend to be computationally expensive. With a rapid expansion of internet and illegal (either due to copyright, privacy infringement it represented or illegal nature of contents), a need for highly efficient methods to find almost exact image matches emerged.

Perceptual hashing algorithms were developed specifically for that need. Introduced by \citet{Fingerprinting1996Schneider}, the main idea is to extract hashes from images, that are based on the visual contents and would be resistant to image modifications that commonly occur on the internet, such as JPEG compression artifacts, resizing or cropping. 

Block mean value hashing algorithms are a fast and performant family of algorithms developed specifically to solve this issue \citep{BlockMean}. They are all built on a simple principle - notably discretization of the image into large "blocks" and an interpretation of such blocks as fully black or fully white based on the average value of pixels within them relative to the average one of the image. The resulting checkerboard pattern is used to build a perceptual hash. The simplest algorithm of the family - \textit{average hash} (Ahash) just uses average values of the block as binary values concatenated to form a hash. \textit{Perceptual hash} (Phash) first converts the image to greyscale and performs a version of a discrete Fourrier transform before applying the checkerboard pattern. Wavelet hash (Whash) does almost the same, but uses a slightly different version of discrete Fourrier transform. The \textit{Difference hash} (Dhash) rather than using average values of pixels in the checkerboard pattern uses differences between average pixel values in neighbouring patterns. Finally the crop resistant hash splits the image in many different segment to still be able to detect small sections that could have been cut out from the image \citep{Crop_res}.

Despite the simplicity, those methods perform surprisingly well on types of images shared on the internet and are used to identify similar or identical images in large-scale analyses of information flow on social networks, such as for instance in \citet{abilov2021voterfraud2020} or in \citet{OnTheOriginsOfMemes2018}, using  Ahash/Whas/Phash and Dhash respectively.

While scalable and well-performing for the detection of image with minimal manipulation, it is impossible to fine-tune them or to force them to focus on specific aspects of images that might be of interest to the researcher.


\subsubsection{Deep Learning Similarity Search}

This shortcoming with regards to detecting novel features at a scale is the one that Deep Learning - based similarity evaluation is meant to address. 

While this class of algorithms has recently come into attention due to Apple's intention to use a version of it - NeuralHash - to detect and report reprehensible images on user devices, with all the privacy, censorship and reliability issues it raises \citep{NeuralHashCollisions2021}, use of deep neural networks for similarity detection is all but new. In \cite{1993LeCunSignatureSiamese}, the authors trained networks to detect signature similarity with a precision sufficient to distinguish authentic ones from fake ones, by using a method they introduced - Siamese Networks. Later the same approach has been applied to learning a similarity metric to recognize face similarity by \cite{FaceSimilarity2005LeCun}, becoming the core principle of the Facebook face matching algorithm.

A major advantage of such approaches is that deep learning networks are able to discover by themselves the robust features of images mapping to what their designers want them to recognize. This learning can occur in self-supervised manner, requiring from their designers to provide them with an architecture and pairs of images they would like the deep neural networks to recognize as either similar (or even identical) or different. This approach is made even stronger by the fact that the deep neural networks recognizing the images as such do not need to be trained from scratch for each application, but rather can be fine-tuned through a process of transfer learning \citep{TransferLearning2012Bengio, TransferLearning1994Caruana}. A neural network already working well could be taught to distinguish even better some features by being fed additional examples of images that the designer wants it to recognize as similar or as different. 

While these capabilities could seem magical, Deep Learning methods in fact do have a direct correspondence to the classical keypoint-based computer vision algorithms. Classical computer vision uses hand-made local feature (keypoints) extractors, that are then aggregated by hand-made algorithms into larger features and finally hand-made methods are used to evaluate the similarity of images based on those aggregate features. Deep Neural Networks learn all those features at the same time, with first layers learning to detect features a few pixels large and subsequent layers aggregating them into larger and more diverse features, up until the last layers that learn to map those features to decisions for whatever tasks the neural networks was designed for. An excellent example is the visualisation provided by \citet{VisualizingImageNetFilters2014ZeilerFergus}. 

A common method to evaluate a global similarity between images is to use the vector of activations of neurons on the last feature layer in common neural network, such as AlexNet \citep{AlexNet2012}, VGG \citep{VGG2015}, Inception network \citep{Inceptionv3} or ResNet \citep{ResNet}; although sometimes more lightweight architectures such as MobileNets \citep{MobileNets2017} or EfficientNet \citep{EfficientNet} are used to speed up and limit resource usage.

A readily available neural network for perceptual and contents-wise image similarity evaluation is the SimCLR family, introduced by \citet{SimCLRv1} and improved in \citet{SimCLRv2}. This family is based on the ResNet architecture truncated at the last feature layer and trained in a self-supervised manner on ImageNet.

All of them are widely used as well in the study of images conveying information on the internet. For instance the MemeSequencer \citep{memesequencer2020dubey} relies on ResNet and AlexNet for the initial image matching, whereas the Motif Mining relies on VGG and MobileNet \citet{MotifMining2022}. However, so far the use of contrastively trained networks have remained minimal and restricted to author methods, such as for instance in \citet{TechnionCornellReview2022}.







\subsection{Contribution}

In this paper we evaluate how methods from different families perform with regards to the detection of image similarities under different assumptions. 

Specifically, we select a few, well-described and well-explored methods with implementations readily available from each family and evaluate their ability to detect image similarity despite perturbations that can be found in images shared online and modified for evasion from similarity detection methods.

In addition to the method performance, we also evaluate how computationally efficient each method is, in order to provide researchers with an idea of computational power and throughput that would be required for them to approach the question at hand.

We present the evaluation metrics for different type of perturbations and performance separately, allowing the researchers to evaluate themselves which method would likely fit best their computational budget and expected image perturbations.

Finally, we perform the evaluation of our methods on a dataset of in-the-wild memes collected in 2018 on Reddit meme-specific datasets, to evaluate whether our insights on smaller dataset with artificial perturbation scale up to the real-world data.



\section{Related work}

Showing that their algorithms performs better than the prior state of the art is a pre-condition for proposing a novel algorithms in computer vision. More comprehensive reviews, comparing several different algorithms are not a novelty either. The ubiquity of SIFT is due in large part to the landmark review by \citet{SIFT_FTW2005}.

However, computer vision algorithms reviews tend to focus on specific classes of methods or on raw performance alone, not accounting for computational cost of running them at scale. The authors of \cite{FeatureExtractionSurvey2021} provide an in-depth and comprehensive review of local keypoint extraction, feature composition and feature matching methods. They do not however cover global block perceptual hashing or integrated deep learning pipelines and benchmark the computational performance only for a fraction of methods. They don't perform a parameters sweep for methods that allow it.
\citet{HPatches2017} focuses as well on local feature matching methods and leaves out global block perceptual hashes as well as integrated deep learning pipelines. They however note contradictions for different methods performance in prior reviews due to parameter selection and perform parameter sweep, as well as report the performance for different methods consistently.
In \citet{2021LightBinaryDescriptorsReview}, the authors review keypoint-based methods while focusing on their computational cost for use in small-scale robotics. In addition to be focused on only one class of methods, the review focuses on a problem that is radically different from the one posed by reuse and remixing of images to convince audiences online; specifically viewpoint change in 3D environment.

On the other hand, recent pipelines designed specifically for analysis of images have benchmarked a small selection of methods, either to propose an own method or to select a global matching method for the first stage of their image analysis pipeline

In \citet{memesequencer2020dubey}, the authors evaluate AlexNet and ResNet in their ability to match the meme template images as a first stage in the pipeline they have proposed. In \citet{TechnionCornellReview2022}, the authors review VGG-16 Deep Neural Network and ORB Local keypoint matching algorithm before proposing their own method, based on a Deep Neural Network trained by contrastive learning. Finally, in \citet{MotifMining2022}, the authors compare Phash, VGG, MobileNet and SURF and a subset of their combinations as a means to compute perceptual similarity of images found in real-life scenarios. Unfortunately, the authors did not analyze relative performance of methods, sweep for parameters or resilience to differnet types of perturbations.

To our knowledge, our review is the first one that covers:
\begin{itemize}
  \item The three classes of algorithms to evaluate the global perceptual similarity of images
  \item Performs a hyperparamter sweep on methods allowing for it (notably the number of keypoints/features in local keypoint matching algorithms)
  \item Evaluates the computational cost of all the methods
  \item Evaluates the robustness of methods to different classes of perturbations likely to be found in images reused and remixed to convince the audiences of online and social media
\end{itemize}

\section{Setup}


\subsection{Computational setup}
All the methods are evaluated on a workstation equipped with Intel Core i9-9900K (8 cores/16 threads CPU), 64 Gb of RAM clocked at 2666 MHz, 2 TB NVME M.2 SSD and an RTX 3080 graphics cards, running an Ubuntu 20.04 LTS distribution. The evaluations were performed within a Docker container, Docker Community Edition, version 20.10.12. The code used Miniconda version 4.12.0, Python 3.9 with cuda version 11.3.1. A more detailed information is available in the \verb|requirements.txt| found in the code repository.

\subsection{Code location}
All the code used in generating figures can be found in the GitHub repository \url{https://github.com/Cyrilvallez/Image-manipulation-detection}, with the commit number \verb|2719616| corresponding to figures generated here.

\subsection{Data}

All images used for evaluation are publicly available. Links for downloading them are provided in the \verb|README.md| in the Github repository. For the synthetic images we created, code is provided to generate the exact same attacks. Moreover, we provide results from our experiments (as json files) and code as well as examples on how to recreate figures from this data. Finally, each experiment we made is documented with enough information (with a yml file) for people to re-run the same one, if one would wish to reproduce results.

\section{Methods}

From the three classes of algorithms described above, we review the following methods :

\subsection{Block perceptual hashing algorithms}

This class of methods was designed specifically for computing speed. This relies on fast binary hash computation, and efficient comparison using the Hamming distance (number of different bits), or equivalently the Bit Error Rate (BER), defined as \begin{equation}
    \textnormal{BER}(a,b) = \frac{H(a,b)}{n}
\end{equation}
where $H(a,b)$ denotes the Hamming distance between two images $a$ and $b$, and $n$ is the number of bits the hashes are made off. The smaller is BER, the higher is the probability that the images look identical.

In our study, we consider the following algorithms :

\begin{itemize}
    \item \textbf{Ahash} Average hash, basically a simple implementation of the block mean value algorithm.
    \item \textbf{Phash} Perceptual hash, an algorithm similar to Ahash, but based on Discrete Cosine Transform (DCT) coefficients to extract the lowest frequencies 
    \item \textbf{Dhash} Difference Hash, computes the gradients between adjacent pixels 
    \item \textbf{Whash} Wavelet hash, similar to Phash but uses Discrete Wavelet Transform (DWT) instead of DCT coefficients
    \item \textbf{Crop res} Crop resistant hash, an implementation of \citet{Crop_res} combining image segmentation and Dhash to extract different hashes by regions in an image.
\end{itemize}

\noindent Note that for crop resistant hash, one has to deal with more than one hash for a single image, and thus must consider a certain number of matching segments for accepting two images as similar. In the following we set it to 1, meaning that if at least one hash from the collection of hashes of two images are similar, then we consider the images as similar. The implementation of these algorithms are taken and adapted from \citet{BuchnerImagehashNodate}.

\subsection{Keypoint extractors and descriptors}

These methods try to detect (extract) a large amount of keypoints in an image, and describe them using binary or non-binary vectors, depending on the approach. In the following study, we restrict the number of features extracted on each image to a relatively small number, for numerical efficiency. The matching between features vectors is done in the same way as for block perceptual hashing algorithms for methods using binary outputs (BER distance), and with Euclidean (L2) distance for non-binary vectors. Since we are interested in image matching rather than feature matching in images, we proceed in the same way as for crop resistant hash, and consider two images as similar if more than a given number of features have a sufficiently small distance between them. In practice, we set the limit to only 1 feature that should match between two images. We consider the following algorithms :

\begin{itemize}
    \item \textbf{SIFT} The SIFT \citep{SIFT} keypoint extractor and descriptor. The Euclidean (L2) distance is used for feature matching.
    \item \textbf{ORB} The ORB \citep{ORB} keypoint extractor and descriptor. The BER distance is used for feature matching.
    \item \textbf{DAISY} The FAST keypoint extractor, coupled with DAISY \citep{DAISY} descriptor. The Euclidean (L2) distance is used for feature matching.
    \item \textbf{LATCH} The FAST keypoint extractor, coupled with LATCH \citep{LATCH} descriptor. The BER distance is used for feature matching.
\end{itemize}

\noindent The implementations of these algorithms are taken from opencv\footnote{  \href{https://docs.opencv.org/4.x/d5/d51/group__features2d__main.html}{https://docs.opencv.org/4.x/d5/d51/group\_\_features2d\_\_main.html}}.

\subsection{Deep neural algorithms}

Finally, we describe as deep neural algorithms the feature extraction process from a neural network. We explore feature extraction from deep CNN that were trained for image classification, as well as models trained in a contrastive fashion. Models trained in such a way naturally learn to be resistant to image modification, which is the subject of our study. 
In order to extract features, we remove the final layers of the models, which were used for a specific task. For image matching, different distance metrics are explored, namely Cosine distance, Euclidean (L1 and L2) distances, and Jensen-Shannon divergence (interpreting the outputs as probabilities to activate or not some parts of the network). Moreover, during inference of these models, all images are pre-processed in the same way as the network was originally trained. All networks are pretrained on ImageNet (ILSVRC2012). We consider the following networks :

\begin{itemize}
    \item \textbf{Inception v3} Inception v3 model \citep{Inceptionv3} pretrained for image classification. 
    \item \textbf{ResNet} ResNets models \citep{ResNet}, with different depth and width multipliers. Pretrained for image classification.
    \item \textbf{EfficientNet B7} EfficientNet B7 model \citep{EfficientNet} pretrained for image classification.
    \item \textbf{SimCLR v1} SimCLR v1 \citep{SimCLRv1} ResNet50 models, with different width mutiplier. Pretrained in a contrastive self-supervised fashion.
    \item \textbf{SimCLR v2} SimCLR v2 \citep{SimCLRv2} ResNets models, with different depth and width mutipliers. Pretrained in a contrastive self-supervised fashion.
    
\end{itemize}

\subsection{Evaluation process}

\subsubsection{Synthetic data}

\begin{figure} 
    \centering
    \subfloat[original]{\includegraphics[width=0.24\linewidth]{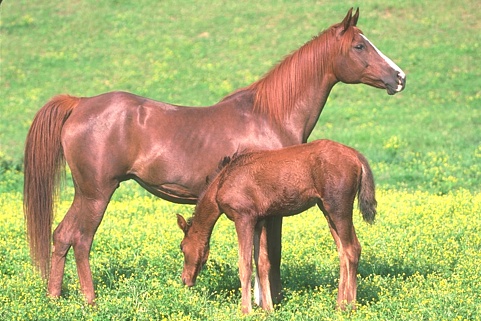}} \,
    \subfloat[contrast 2]{\includegraphics[width=0.24\linewidth]{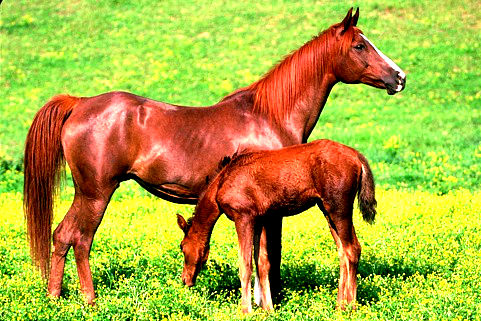}}\,
    \subfloat[median filter 7x7]{\includegraphics[width=0.24\linewidth]{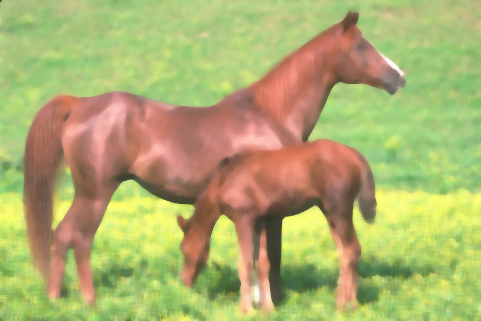}} \,
    \subfloat[s\&p noise 0.15]{\includegraphics[width=0.24\linewidth]{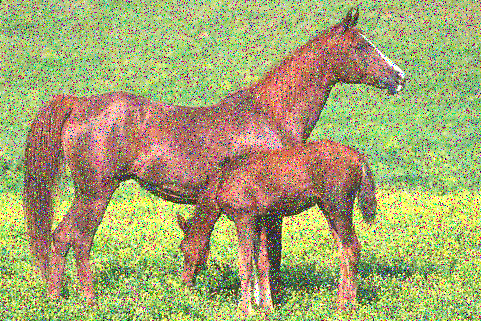}} \\
    \subfloat[cropping 60]{\includegraphics[width=0.24\linewidth]{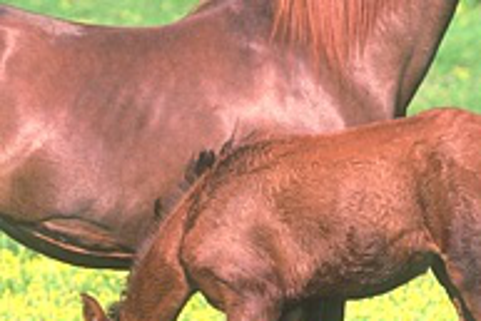}} \,
    \subfloat[rotation 60]{\includegraphics[width=0.24\linewidth]{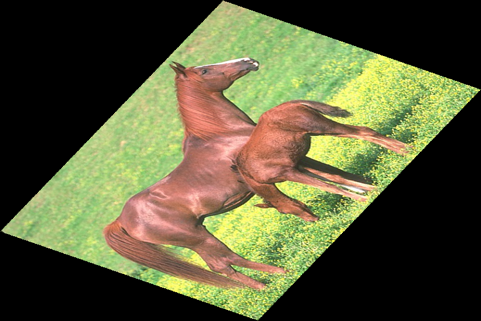}} \,
    \subfloat[shearing 20]{\includegraphics[width=0.24\linewidth]{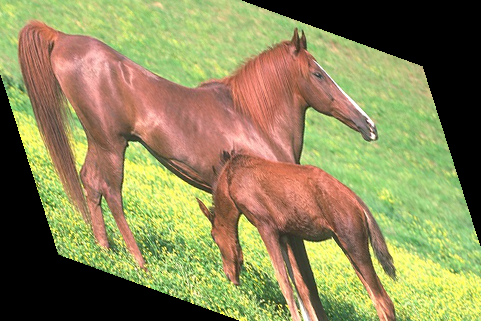}} \,
    \subfloat[text length 50]{\includegraphics[width=0.24\linewidth]{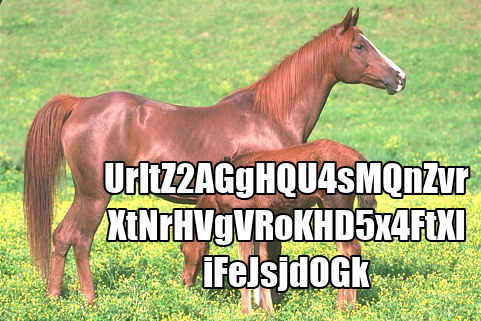}}
    \caption{Illustration of some of the artificial perturbations applied}
    \label{fig:artificial_perturbations}
\end{figure}

To evaluate the performances of all these algorithms, we use two datasets : the BSDS500 dataset (introduced in \citet{BSDS500}) containing 500 images, and the evaluation set of ImageNet (ILSVRC2012 validation set) containing 50,000 images. For both datasets, we first randomly split the images in two equal parts, creating what we call the "experimental" and "control" groups. The experimental set is used as the database of images to identify. We then randomly sample a given number of images in both sets (100 in the following). On those sampled images, we apply a set of artificial perturbations described in Table \ref{tab:perturbations_table} (see Figure \ref{fig:artificial_perturbations} for visual representation of some attacks). Total of 58 perturbations are applied on each image, giving 11,600 perturbed images. Images sampled from the experimental set are used to check for correct matches/incorrect misses, while image from the control set act as proxy to evaluate incorrect matches/correct misses. We evaluate performances with regards to recall and false positive rate (FPR), and most importantly we are interested in the Receiver Operating Characteristic (ROC) curves. Such curves are obtained by varying the threshold in the distance metric described above such that we consider two images as identical. Explicitly, we consider two images as identical if 

\begin{equation}
    d(a,b) \leq \eta
\end{equation}

where $d(a,b)$ is the distance metric between two images $a$ and $b$ (for example BER distance), and $\eta$ is the threshold. The higher $\eta$ is, the more likely we are to obtain false positive matches.

\begin{table}[ht]
\centering
\begin{tabular}{|ll|l|}
\hline
\multicolumn{2}{|c|}{Distortion types}           & \multicolumn{1}{c|}{Parameters}  \\ \hline
\multicolumn{1}{|l|}{\multirow{6}{*}{Noise-like}}  & Gaussian noise       & Variance (0.01, 0.02, 0.05)    \\
\multicolumn{1}{|l|}{} & Speckle noise           & Variance (0.01, 0.02, 0.05)      \\
\multicolumn{1}{|l|}{} & Salt \& Pepper noise    & Amount (0.05, 0.1, 0.15)         \\
\multicolumn{1}{|l|}{} & Gaussian filter         & Kernel size (3x3, 5x5, 7x7)      \\
\multicolumn{1}{|l|}{} & Median filter           & Kernel size (3x3, 5x5, 7x7)      \\
\multicolumn{1}{|l|}{} & Jpeg compression        & Quality factor (10, 50, 90)      \\ \hline
\multicolumn{1}{|l|}{\multirow{5}{*}{Geometric}}   & Cropping + rescaling & Percentage (5, 10, 20, 40, 60) \\
\multicolumn{1}{|l|}{} & Rotation + rescaling    & Degrees (5, 10, 20, 40, 60)      \\
\multicolumn{1}{|l|}{} & Shearing                & Degrees (1, 2, 5, 10, 20)        \\
\multicolumn{1}{|l|}{} & Scaling                 & Ratio (0.4, 0.8, 1.2, 1.6)       \\
\multicolumn{1}{|l|}{} & Addition of text        & Text length (10, 20, 30, 40, 50) \\ \hline
\multicolumn{1}{|l|}{\multirow{4}{*}{Enhancement}} & Color modification   & Factor (1/2, 2/3, 3/2, 2)    \\
\multicolumn{1}{|l|}{} & Sharpness modification  & Factor (1/2, 2/3, 3/2, 2)      \\
\multicolumn{1}{|l|}{} & Contrast modification   & Factor (1/2, 2/3, 3/2, 2)      \\
\multicolumn{1}{|l|}{} & Brightness modification & Factor (1/2, 2/3, 3/2, 2)      \\ \hline
\end{tabular}
\vspace{0.3cm}
\caption{Artificial perturbations description}
\label{tab:perturbations_table}
\end{table}

\subsubsection{In-the-wild memes dataset}
\begin{figure} 
    \centering
    \subfloat[template]{\includegraphics[width=0.24\linewidth]{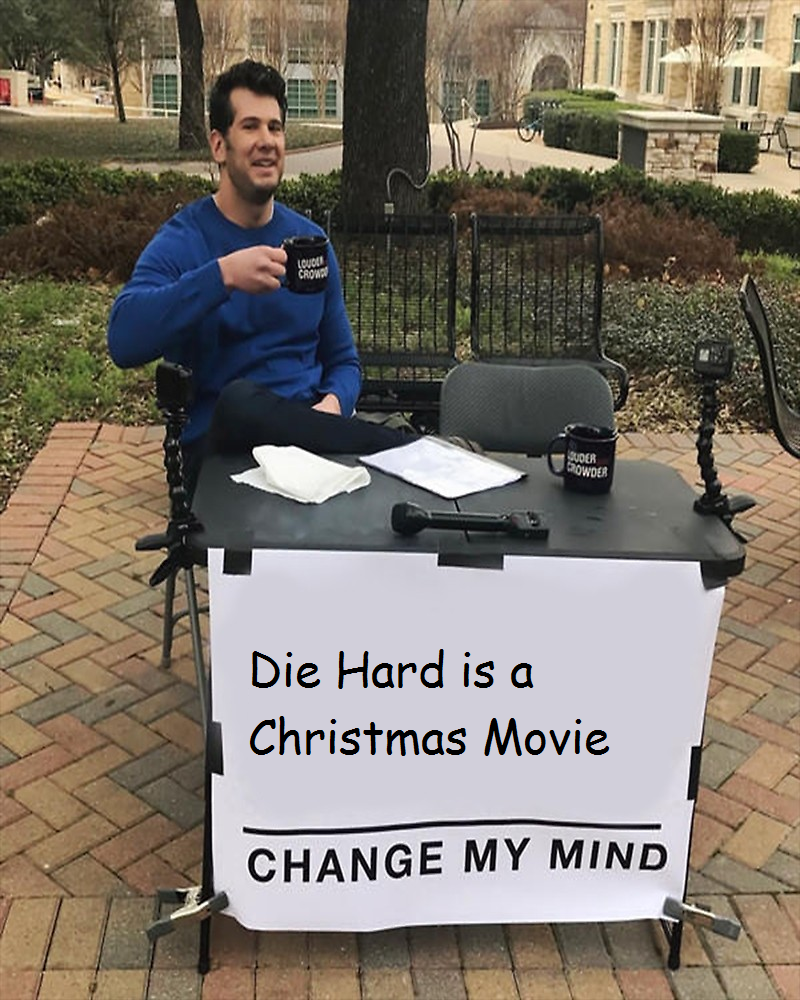}} \,
    \subfloat[]{\includegraphics[width=0.24\linewidth]{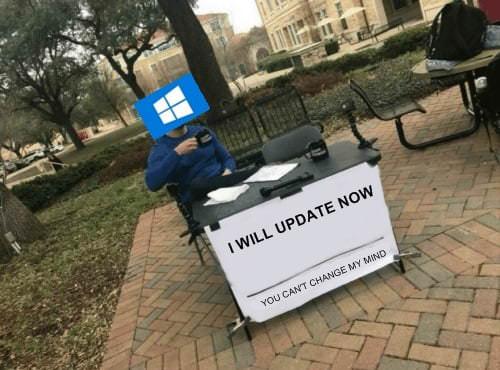}} \,
    \subfloat[]{\includegraphics[width=0.24\linewidth]{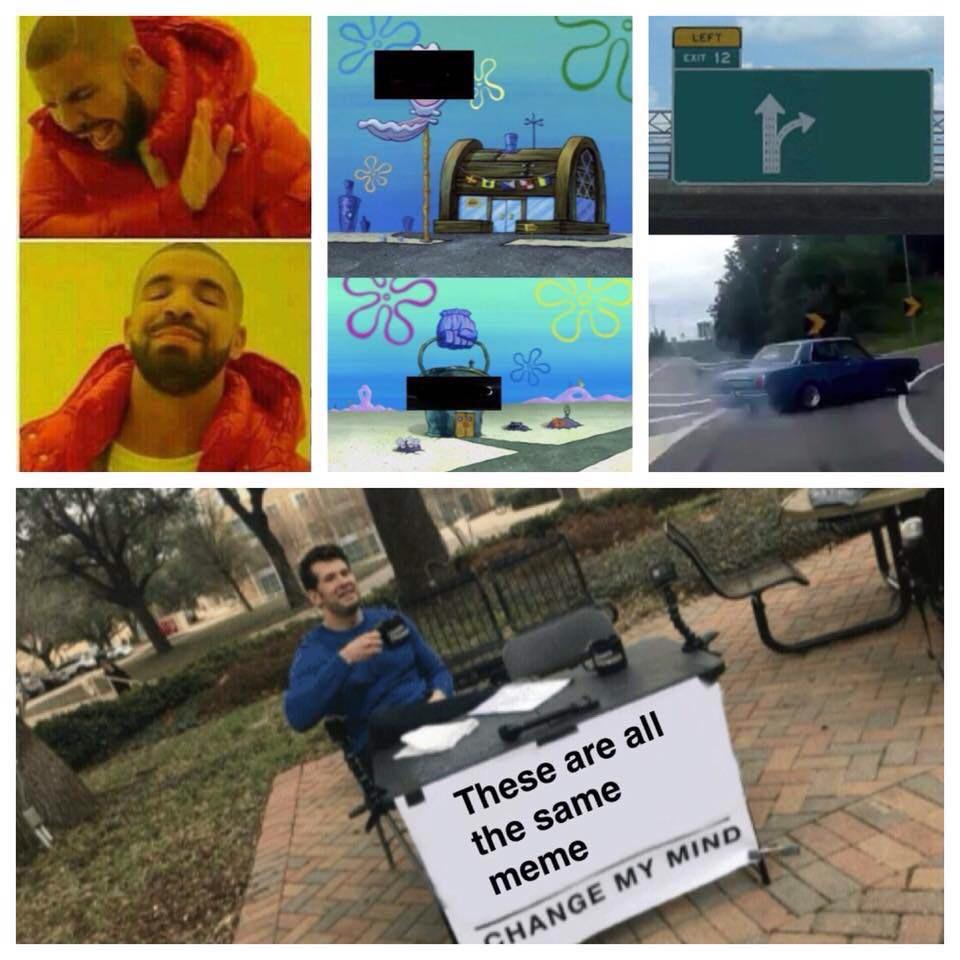}} \,
     \subfloat[]{\includegraphics[width=0.24\linewidth]{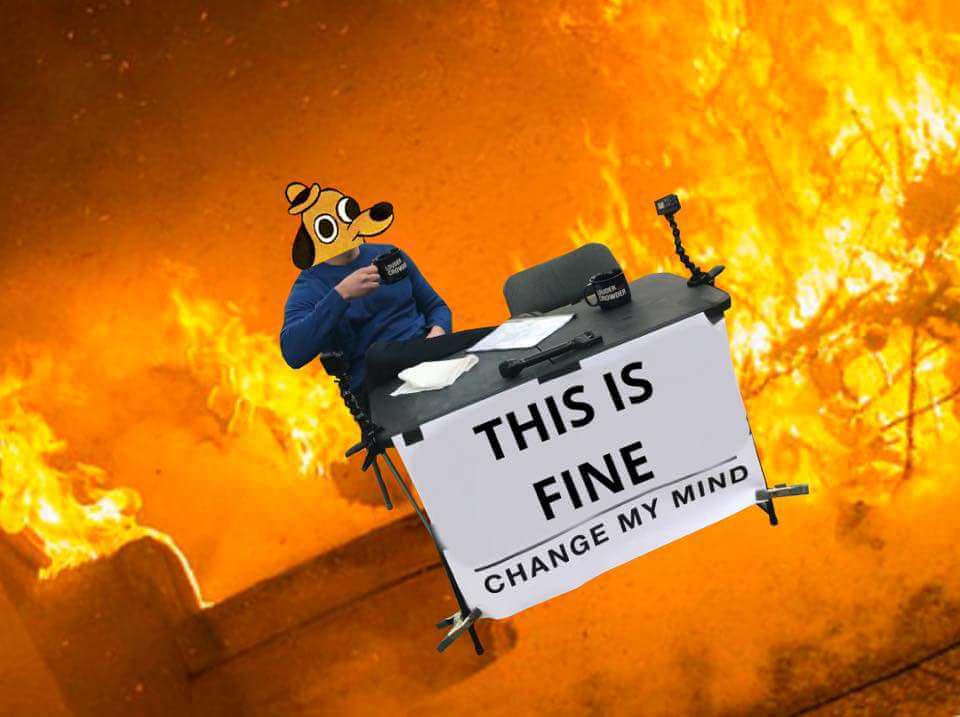}} \\
    \subfloat[template]{\includegraphics[width=0.24\linewidth]{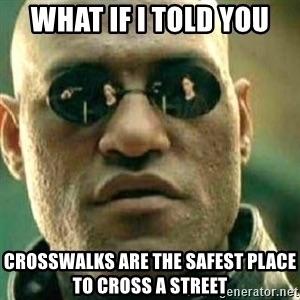}} \,
    \subfloat[]{\includegraphics[width=0.24\linewidth]{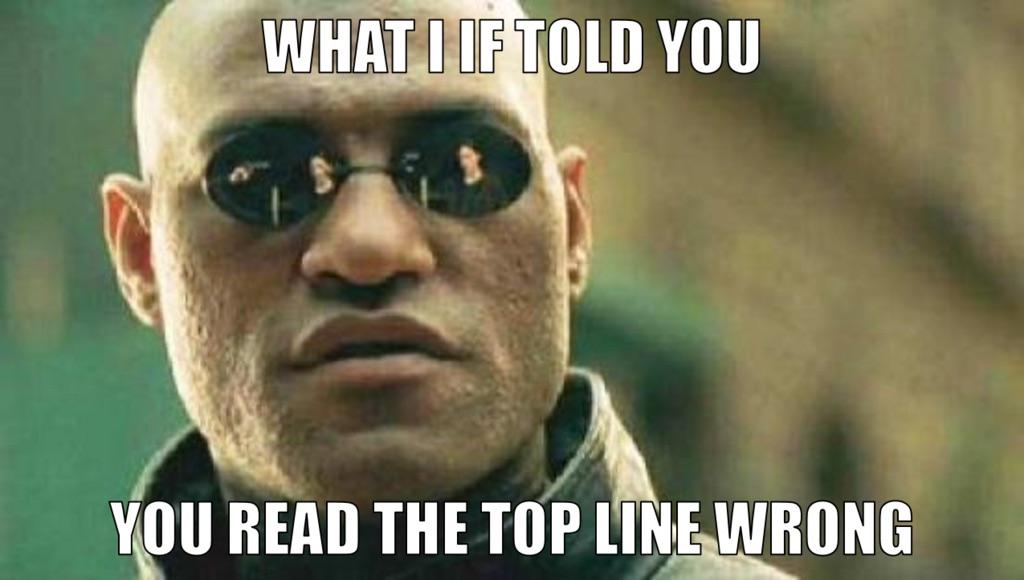}} \,
    \subfloat[]{\includegraphics[width=0.24\linewidth]{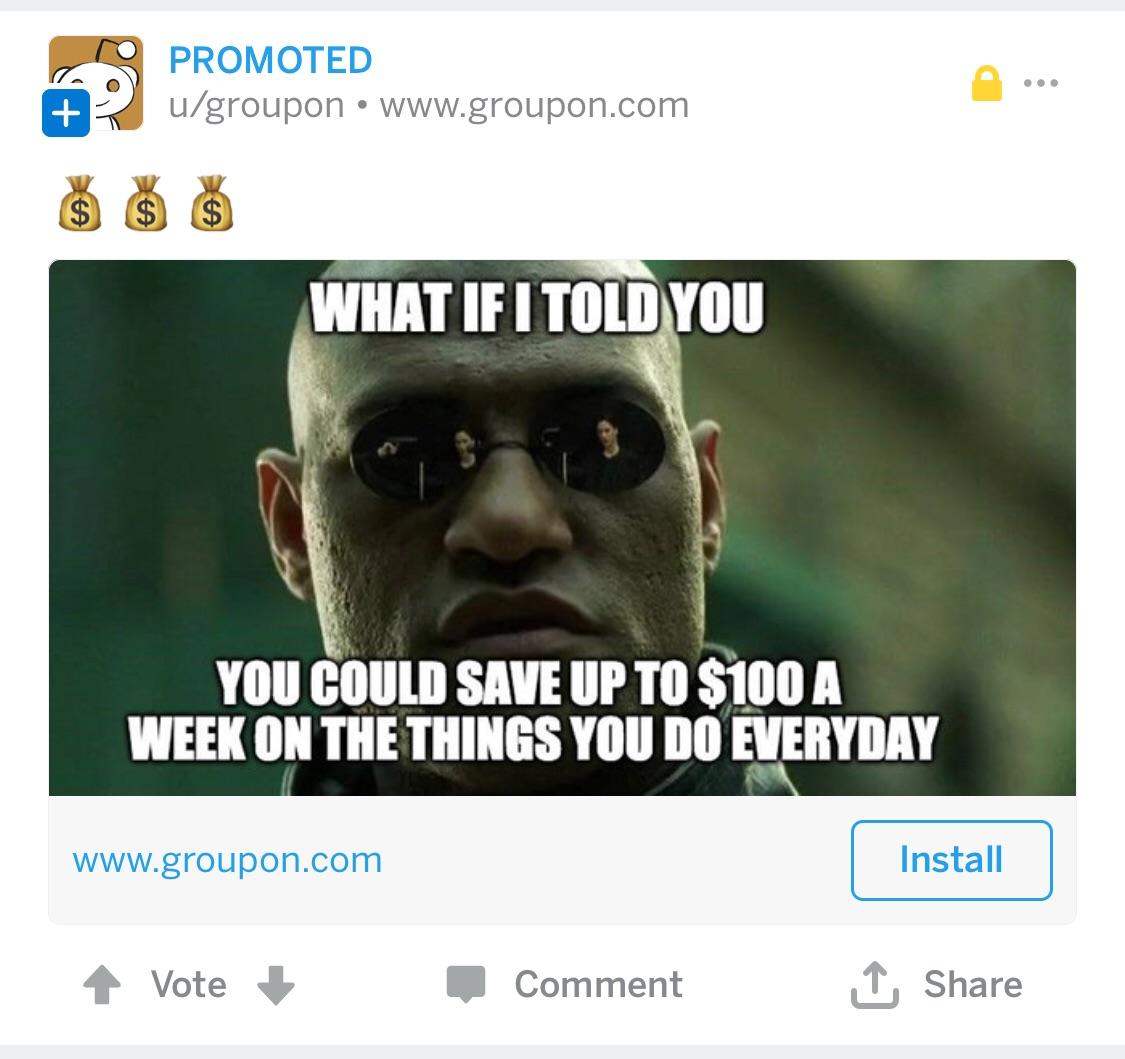}} \,
    \subfloat[]{\includegraphics[width=0.24\linewidth]{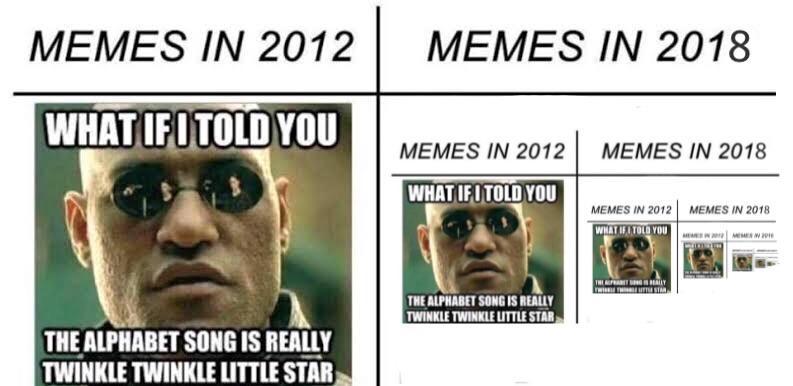}}
    \caption{Illustration of the Kaggle memes dataset. On the left are the templates spanning the memes on the right. While some memes are similar and vary only by size and text, some other are quite different.}
    \label{fig:kaggle_memes}
\end{figure}

Second phase of the evaluation involved using the Kaggle dataset "Most popular memes templates of 2018". This dataset contains 43660 unique images, each manually annotated with one of the 250 corresponding meme templates, supplied in the accompanying "reddit\_posts.csv" metadata file. A single image was randomly chosen to represent the "base" version of the template. Figure \ref{fig:kaggle_memes} illustrates the dataset.

Reddit is a general social network, composed of pages dedicated to different subjects of interests (\textit{sub}), where users submit content and other users within the \textit{sub} vote on the interest of the content and comment on it. It has specifically several \textit{subs} dedicated to submitting and grading memes. The dataset in question collected the most popular memes posted to those \textit{subs} and manually annotated the template used in each of them, providing a ground truth for us to evaluate the detection capabilities of different methods.

Unlike synthetic data this in-the-wild data does not adhere to any assumptions about how memes or templates are used, and hence are a good test for expected in-the-wild performance of different methods.

\section{Results}

In the following, the symbol $*$ before a neural algorithm means SimCLR v1, and $**$ means SimCLR v2. For example, *ResNet50 2x means a ResNet with depth 50 and width multiplier of 2, that was trained using the SimCLR v1 procedure.

\subsection{Choosing the hash length}

\begin{figure} 
    \centering
    \subfloat[AUC]{\includegraphics[width=0.45\linewidth]{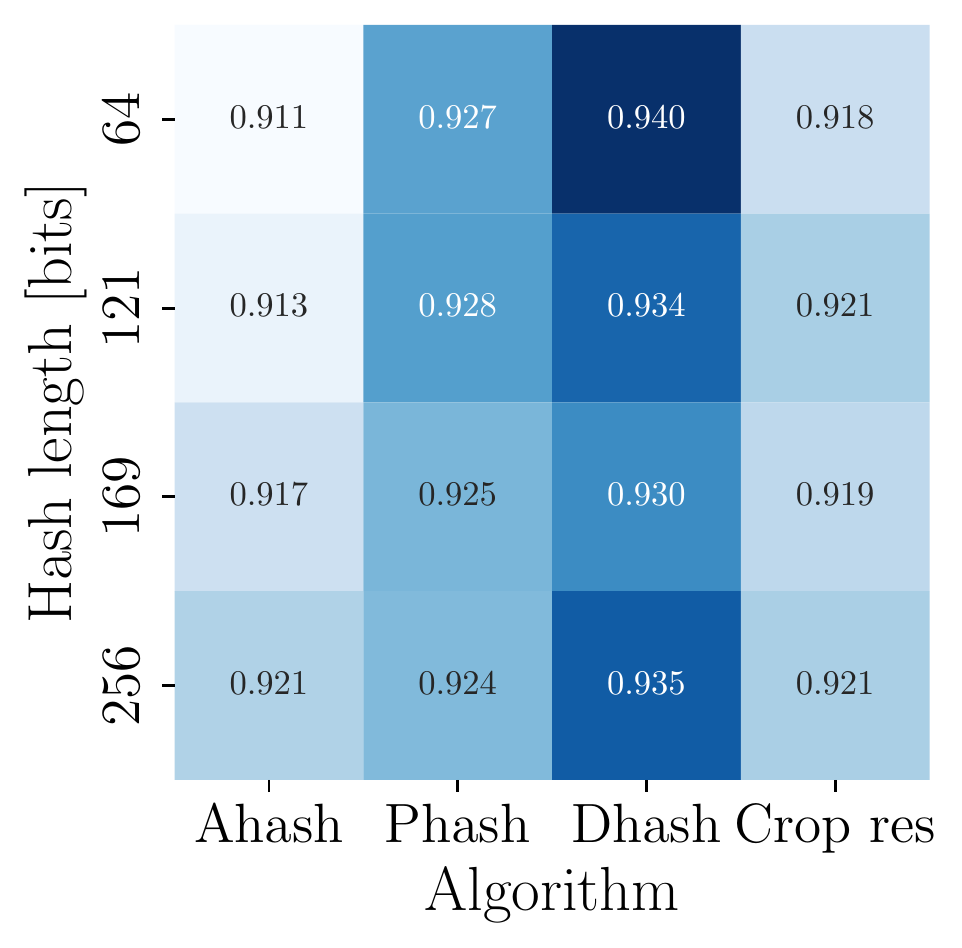}} \quad
    \subfloat[Computation time (min:sec)]{\includegraphics[width=0.45\linewidth]{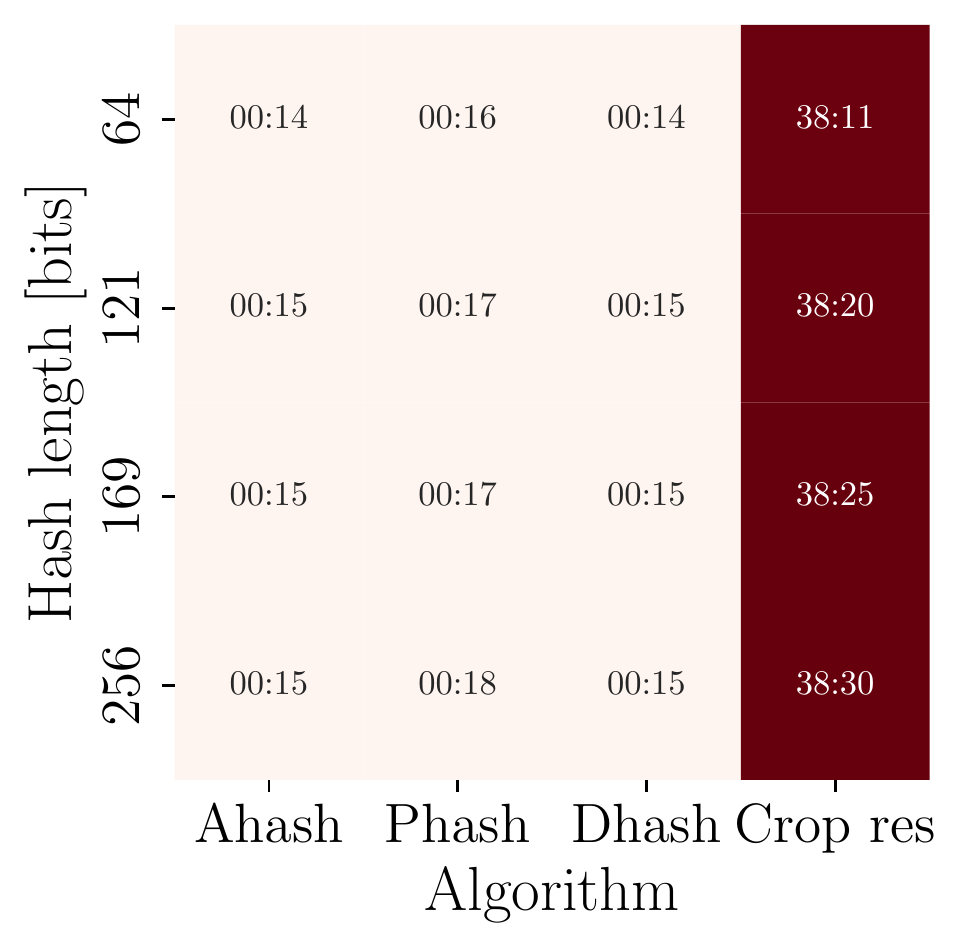}}
    \caption{Area Under the ROC Curve (AUC) for all perturbations and computation time when varying the hash length for block perceptual hashing algorithms. Whash is not included given it does not allow to set an arbitrary hash length.}
    \label{fig:comparison_AUC_classical}
\end{figure}

We are first interested in finding out what hash lengths are best for block mean perceptual hashing algorithms, for which the length can be easily adjusted. Figure \ref{fig:comparison_AUC_classical} show the Area Under the ROC Curve (AUC) for such methods, for different choices of hash lengths. Note that a AUC of 1 means a perfect classifier, while 0.5 is a random guess. In the following we set the length to $64$ as larger values do not seem to improve reliability for each method, and even have a negative impact on the best one, Dhash.

\subsection{Choosing the number of keypoints}

\begin{figure} 
    \centering
    \subfloat[AUC]{\includegraphics[width=0.45\linewidth]{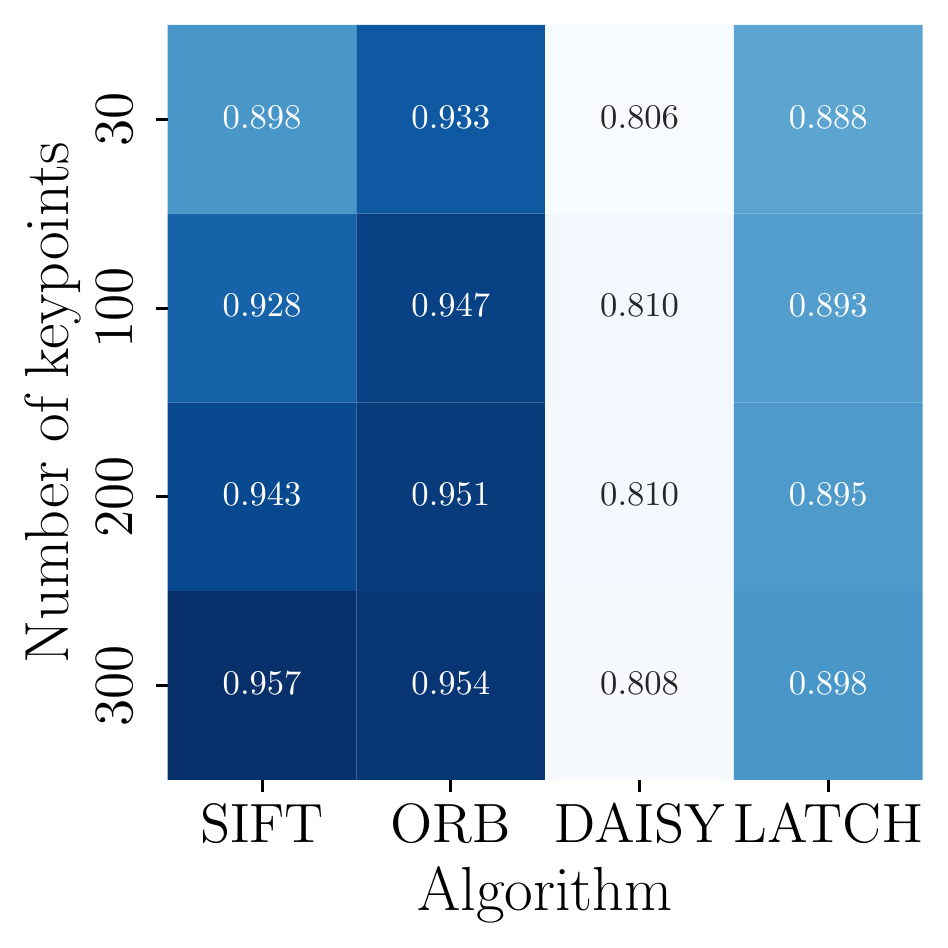}} \quad
    \subfloat[Computation time (min:sec)]{\includegraphics[width=0.45\linewidth]{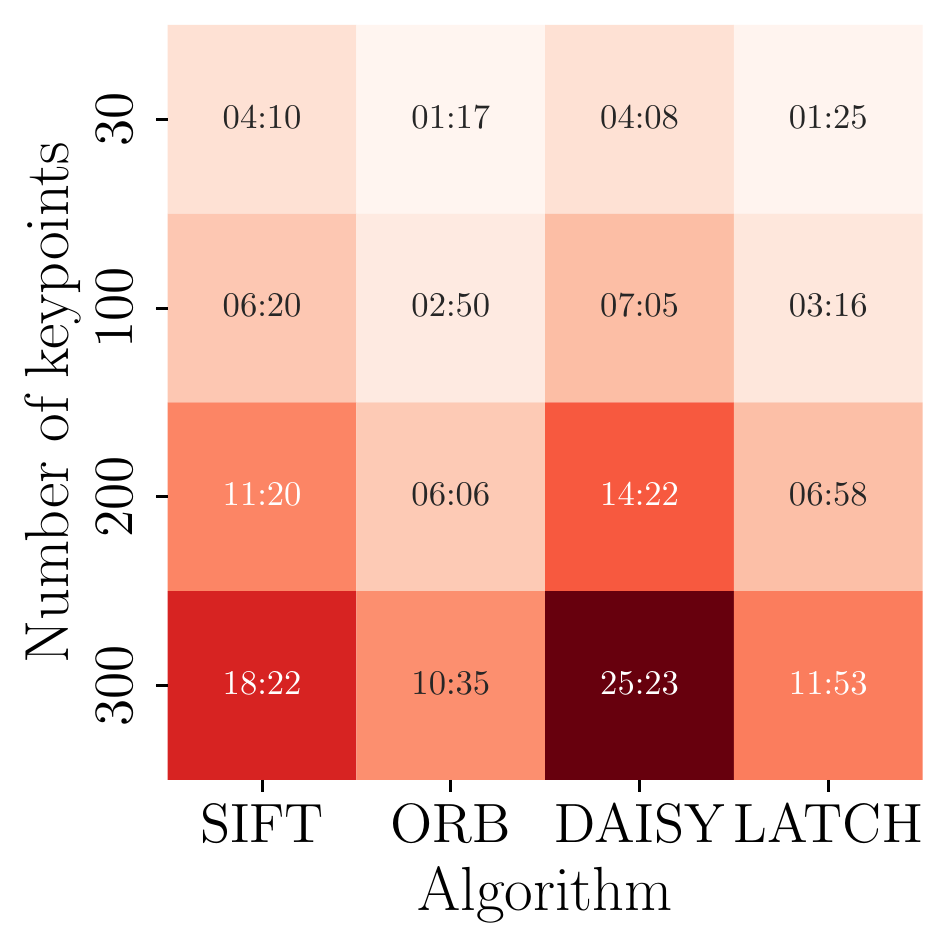}}
    \caption{Area Under the ROC Curve (AUC) for all perturbations and computation time when varying the number of keypoints for keypoint extractors and descriptors.}
    \label{fig:comparison_features}
\end{figure}

In the same manner as the hash length, the number of keypoints extracted from images can be adapted for keypoint extractors. Once again we explore the AUC for different numbers, and also look at the computation time. Figure \ref{fig:comparison_features} shows the results as heatmaps. While the number of keypoints almost does not have any effect on DAISY and LATCH, it does improve SIFT and ORB results. However the computation time quickly increases as we increase the keypoints number, and we thus choose to set it to 30 in the following, as this already gives decent results and is closer to computation time of other methods.

\subsection{Superiority of Jensen-Shannon divergence}

\begin{figure} 
    \centering
    \subfloat[AUC]{\includegraphics[width=0.5\linewidth]{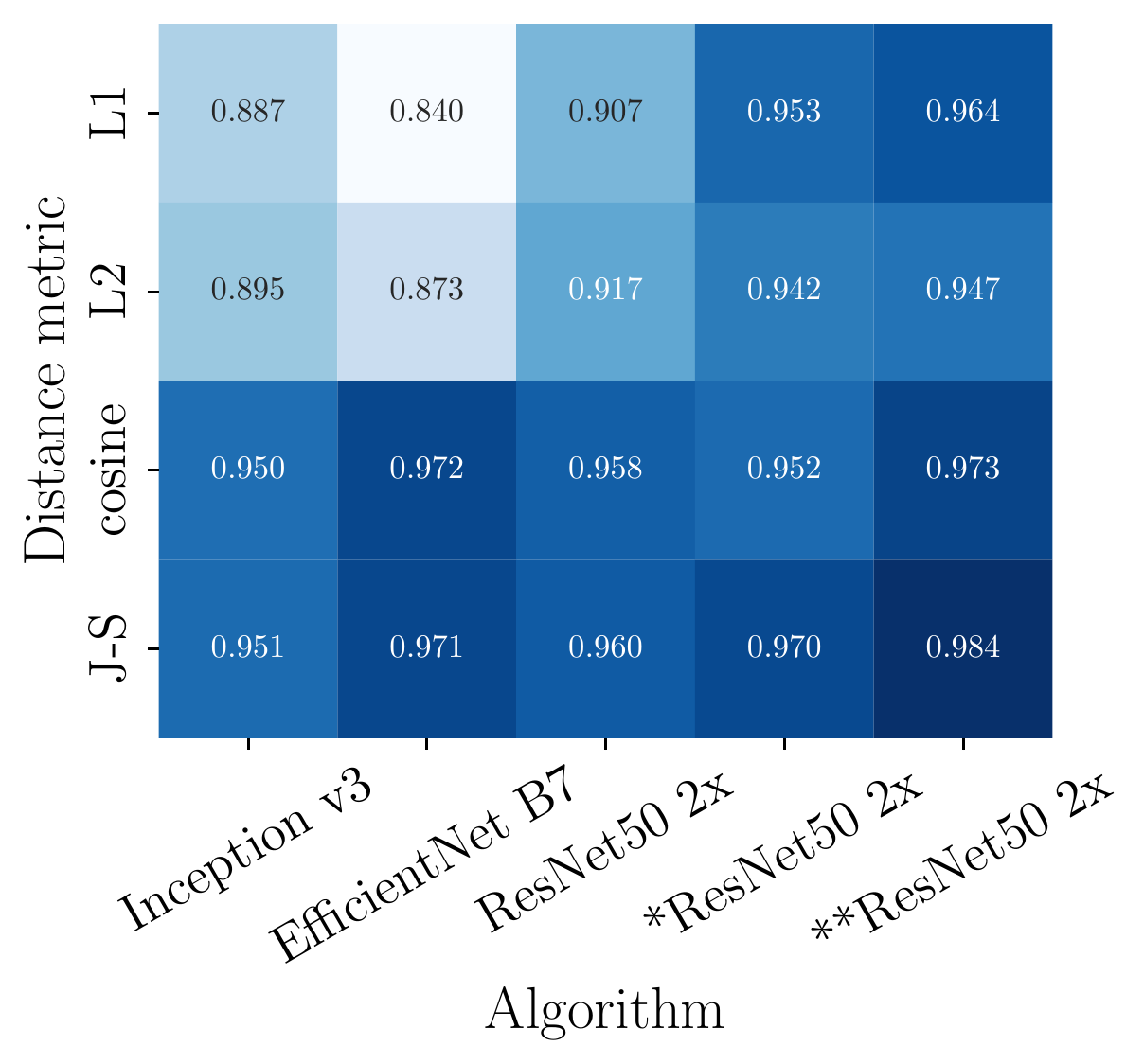}}
    \subfloat[Computation time (min:sec)]{\includegraphics[width=0.5\linewidth]{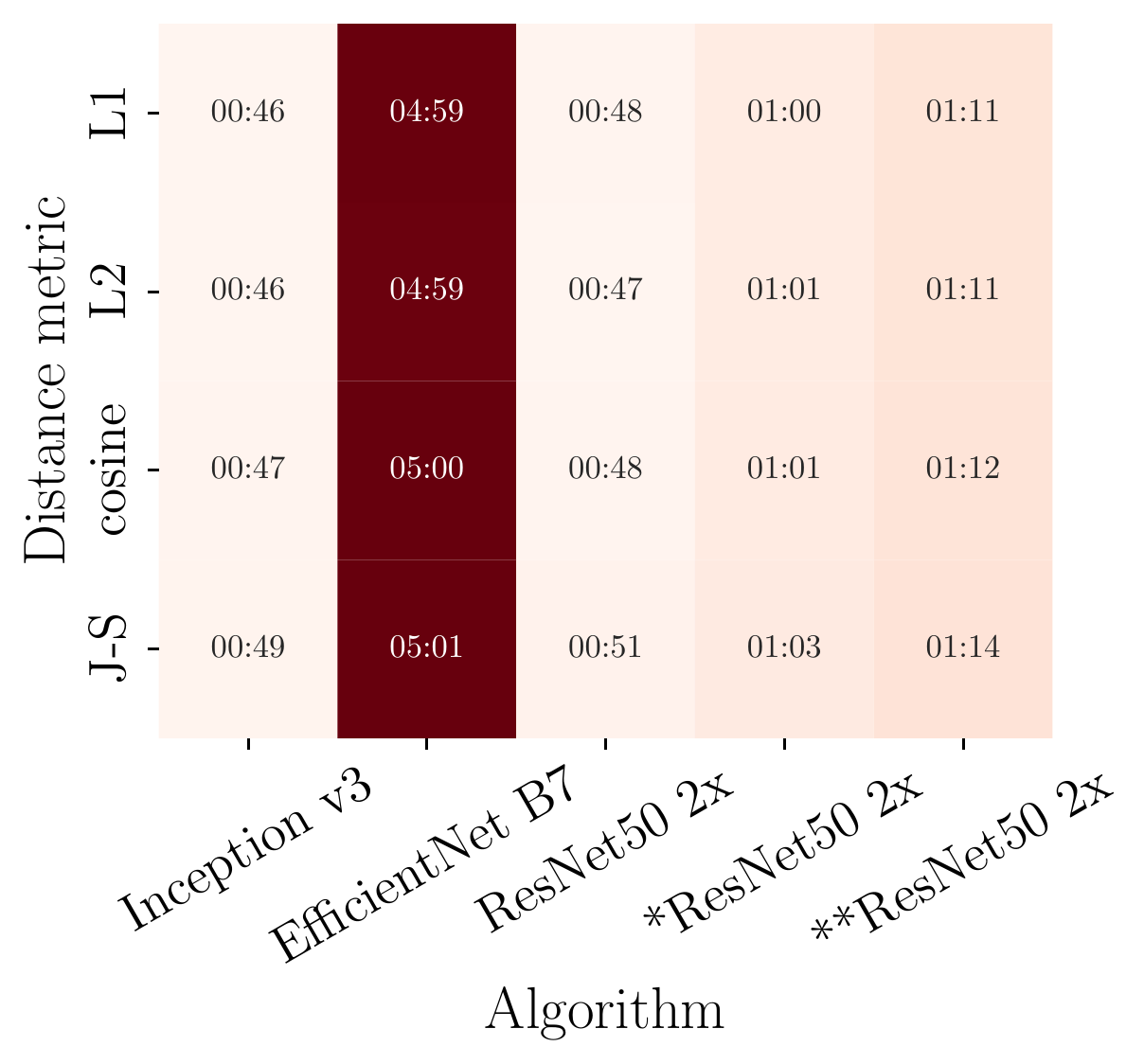}}
    \caption{Area Under the ROC Curve (AUC) for all perturbations and computation time when varying the distance metric for neural algorithms. J-S is Jensen-Shannon divergence.}
    \label{fig:comparison_neural}
\end{figure}

While the output dimensionality of neural algorithms is fixed by the dimensionality of the bottleneck layer we use to represent the image, one can use any distance metric to compare images. In Figure \ref{fig:comparison_neural} we demonstrate the superiority of Jensen-Shannon divergence over L1 norm, L2 norm, and cosine distance. It consistently gives better or equivalent results (the gain being more pronounced for models trained in a constrative way) while incurring only a very small overhead.

\subsection{General performances}

\begin{figure} 
    \centering
    \subfloat[block perceptual hashing algorithms]{\includegraphics[width=0.5\linewidth]{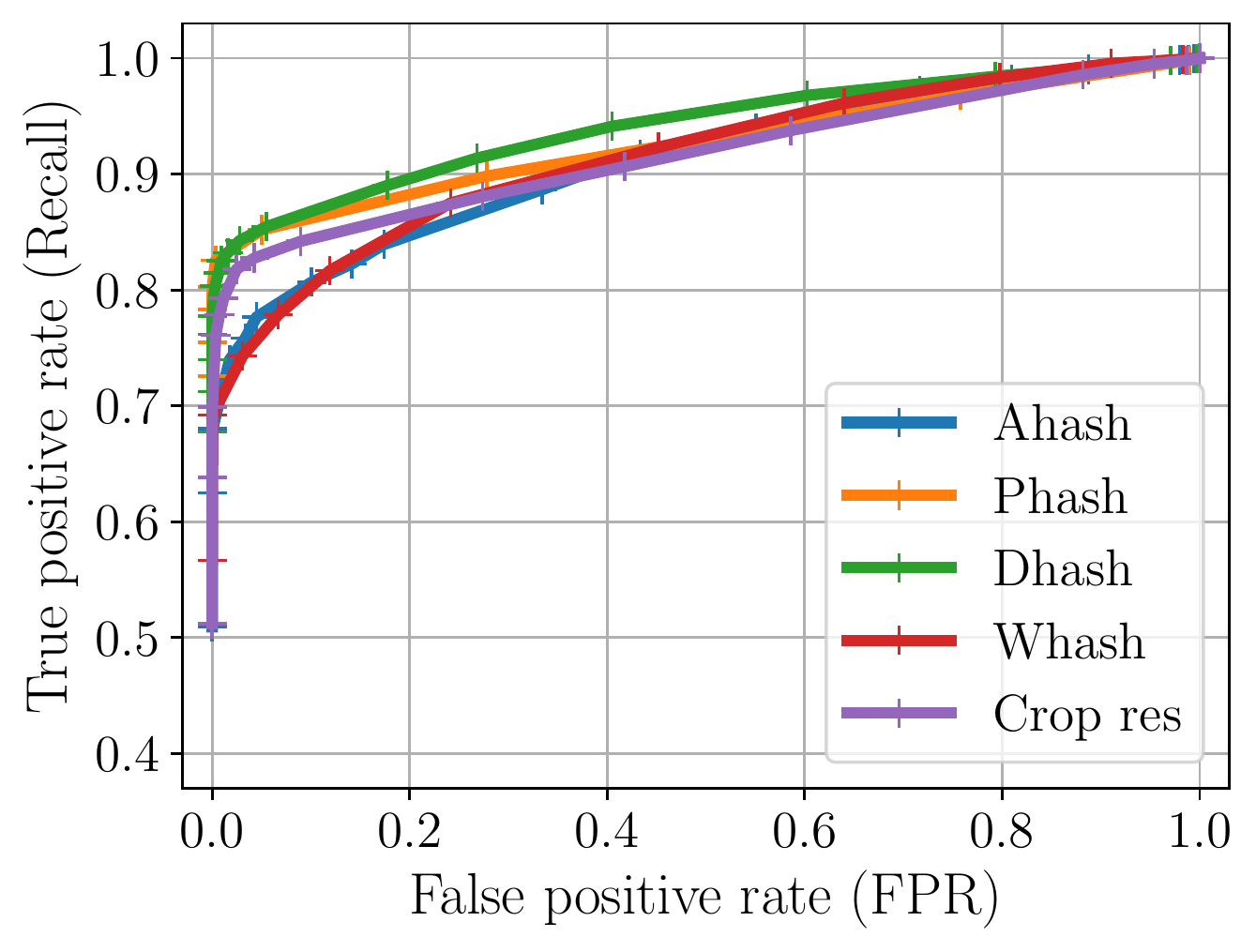}}
    \subfloat[keypoint-based algorithms]{\includegraphics[width=0.5\linewidth]{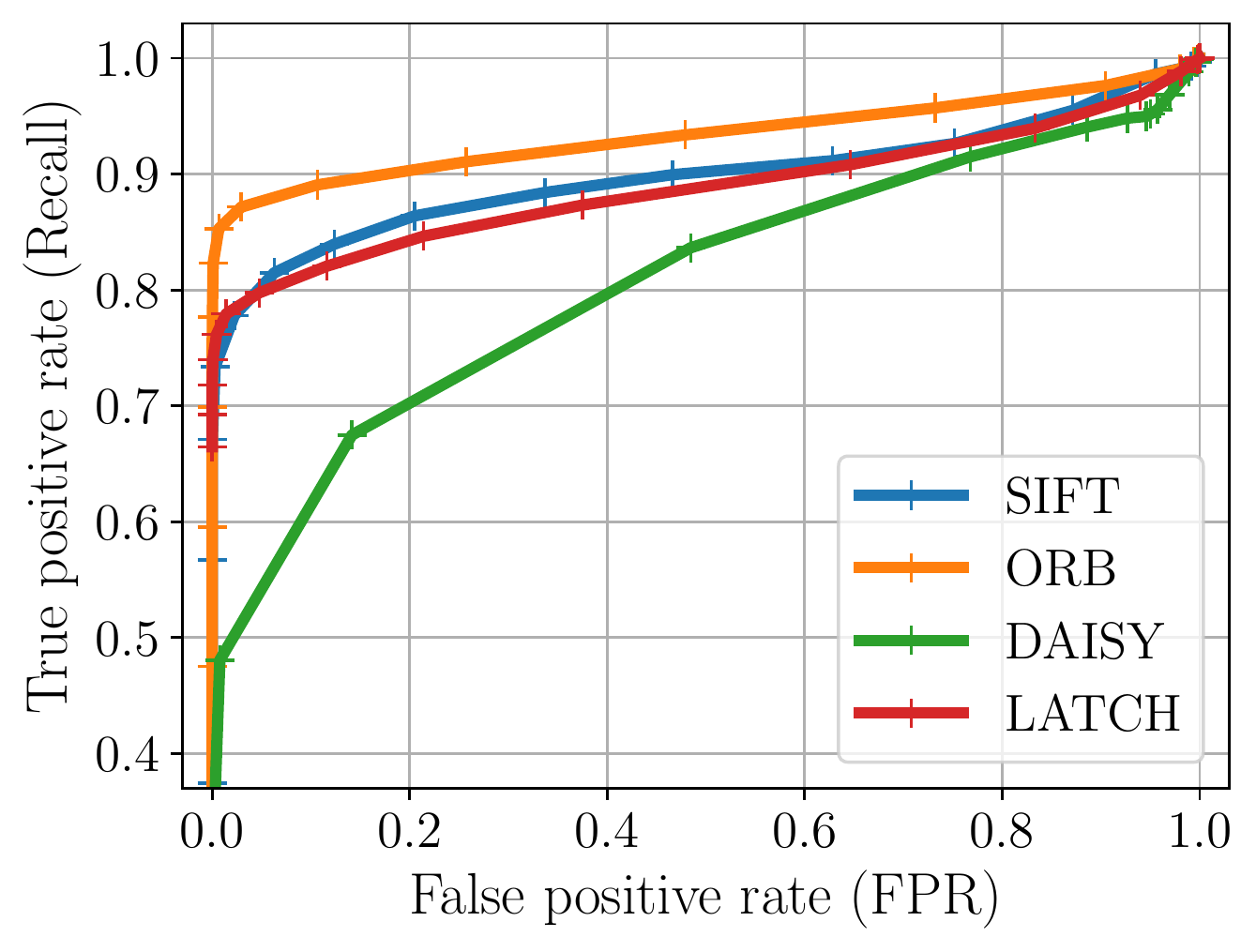}} \\
    \subfloat[neural algorithms trained for \newline classification]{\includegraphics[width=0.5\linewidth]{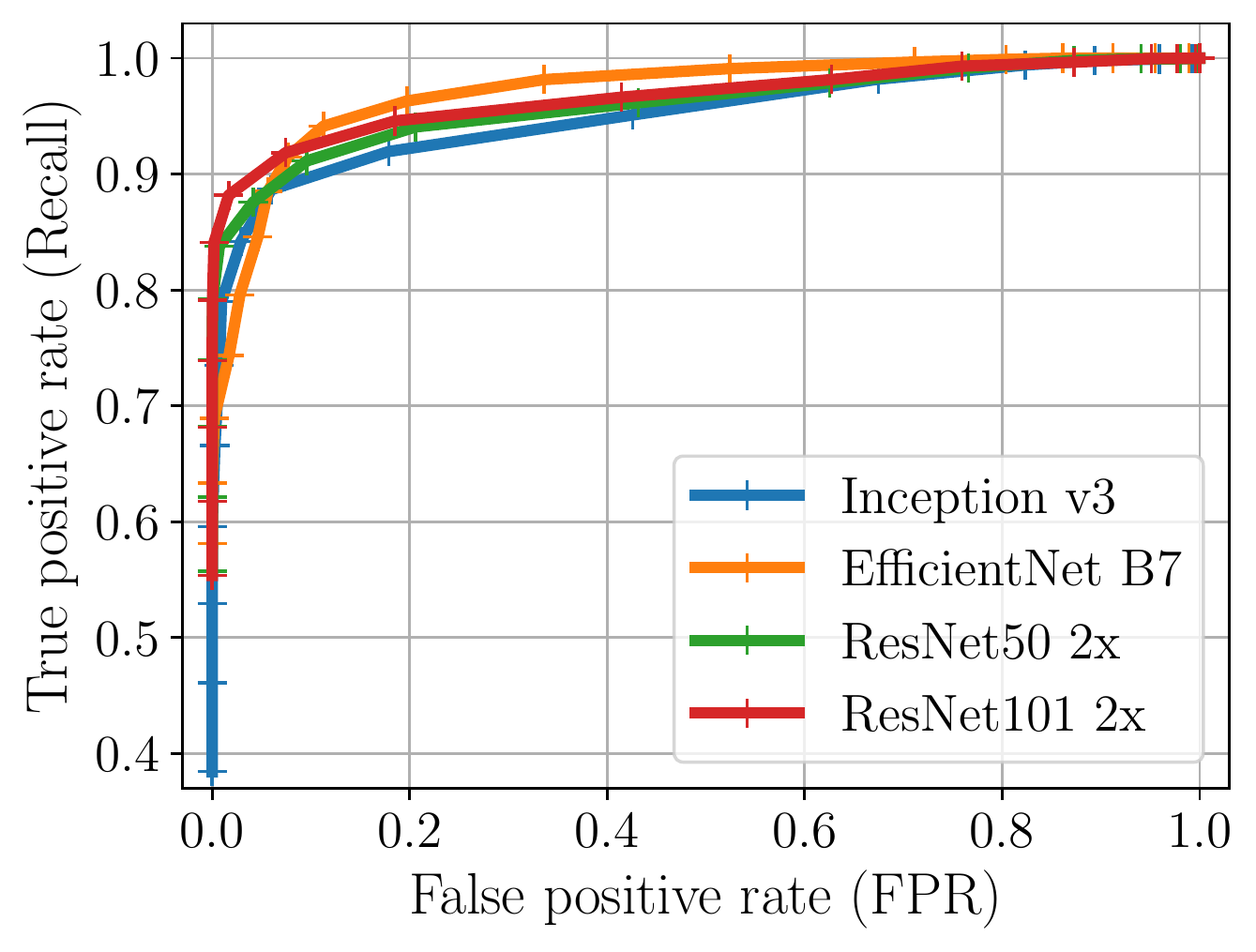}}
    \subfloat[neural algorithms trained in contrastive fashion]{\includegraphics[width=0.5\linewidth]{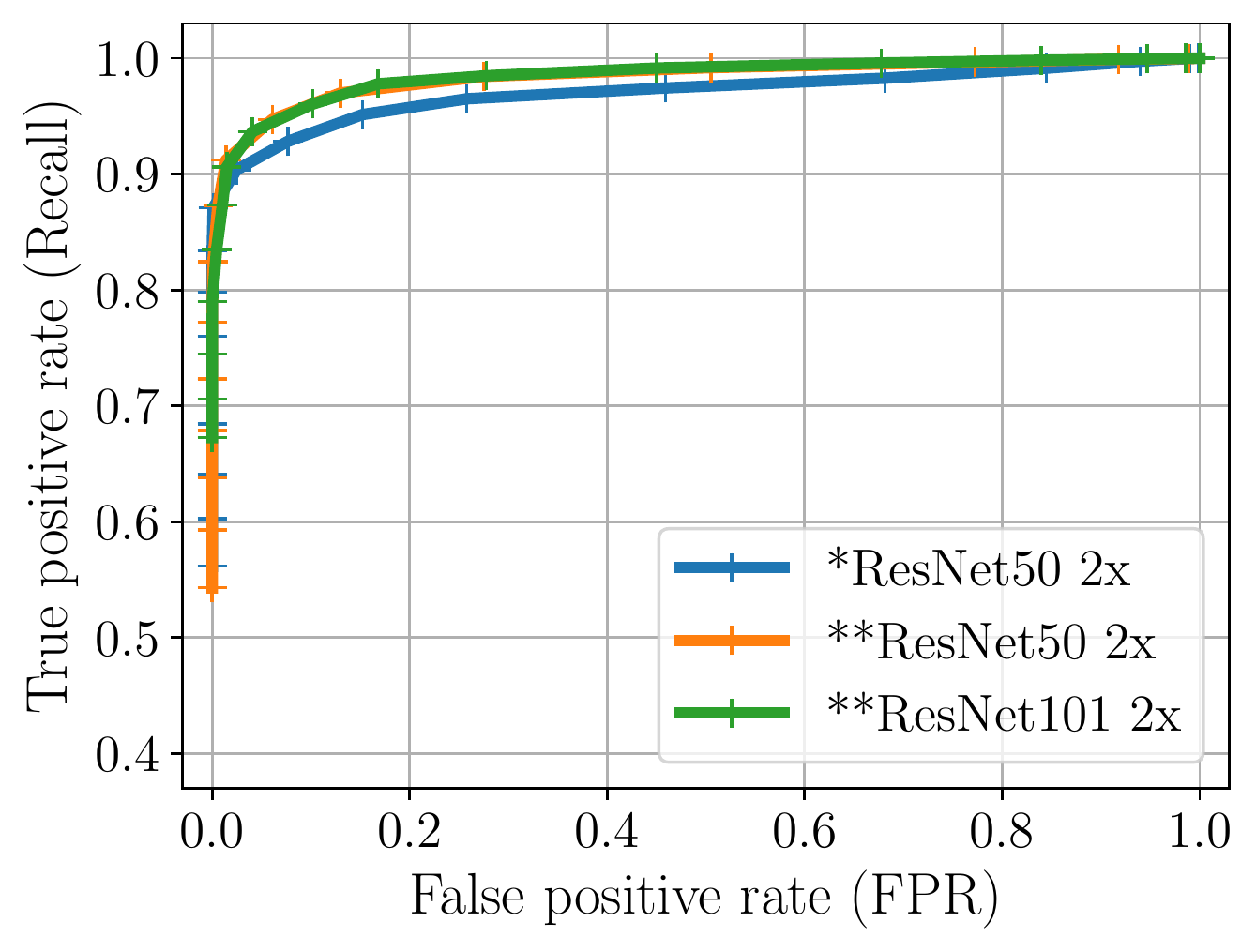}}
    \caption{Overall performances for each algorithms. Block perceptual hashing algorithms were used with hashes of length 64, keypoint-based algorithms were used with 30 keypoints, and neural algorithms with Jensen-Shannon distance.}
    \label{fig:roc_curves_BSDS500}
\end{figure}

\begin{figure} 
    \centering
    \includegraphics[width=0.85\linewidth]{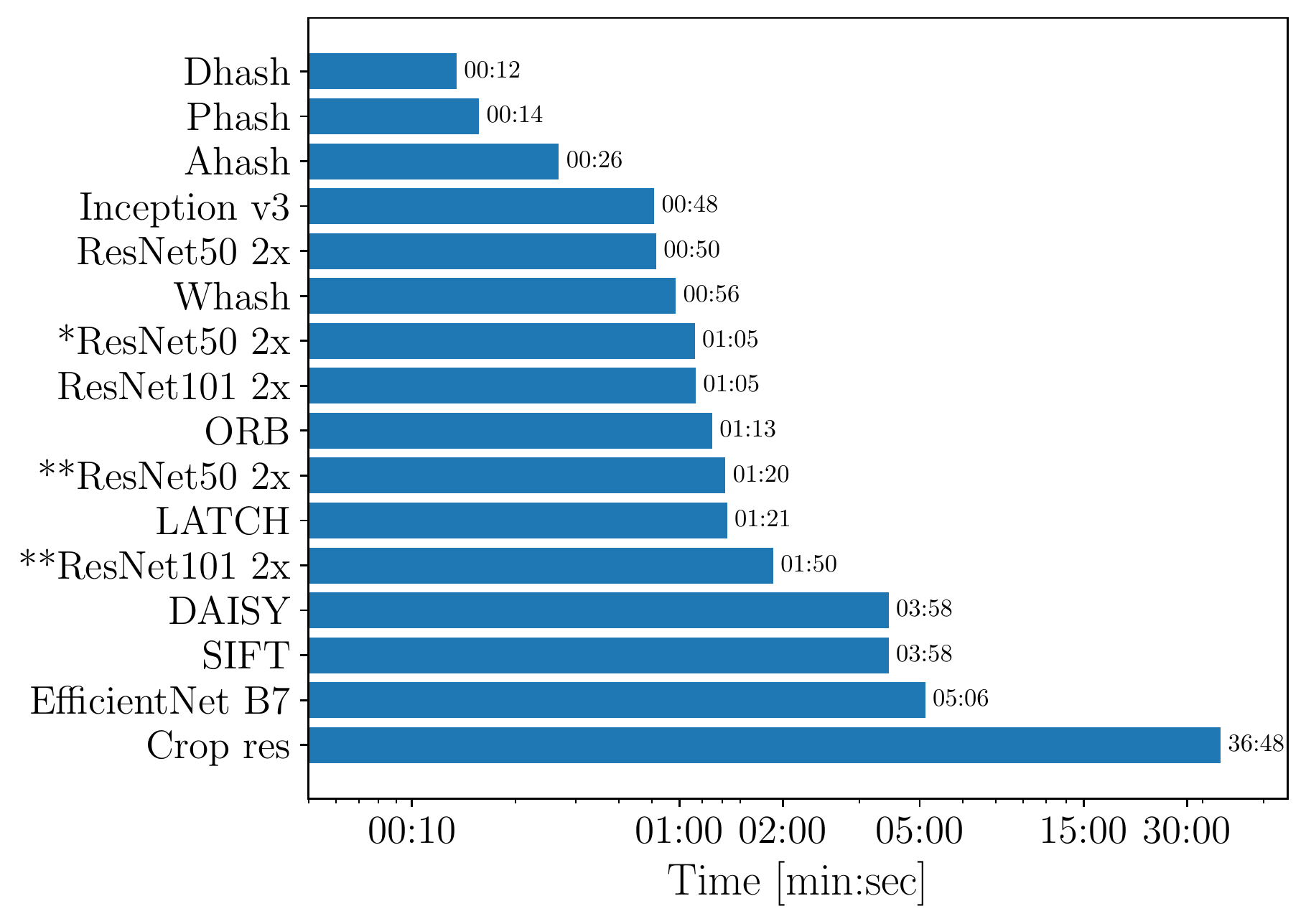}
    \caption{Computation time for each method, in logarithmic scale. This includes the time to create the database, and to perform the hashing and matching process (as well as any preprocessing, mostly for neural methods). Crop resistant hash is much more costly than other methods.}
    \label{fig:time_BSDS500}    
\end{figure}

In Figures \ref{fig:roc_curves_BSDS500} and \ref{fig:time_BSDS500}, we show the complete ROC curves and computational time needed for each algorithm. While block perceptual hashing algorithms are very fast, they do not give results as good as neural networks trained to differentiate between images. Crop res is much more inefficient than all other, and does not perform better.

\subsection{Performances with regard to each attack}

\begin{figure} 
    \centering
    \includegraphics[width=\linewidth]{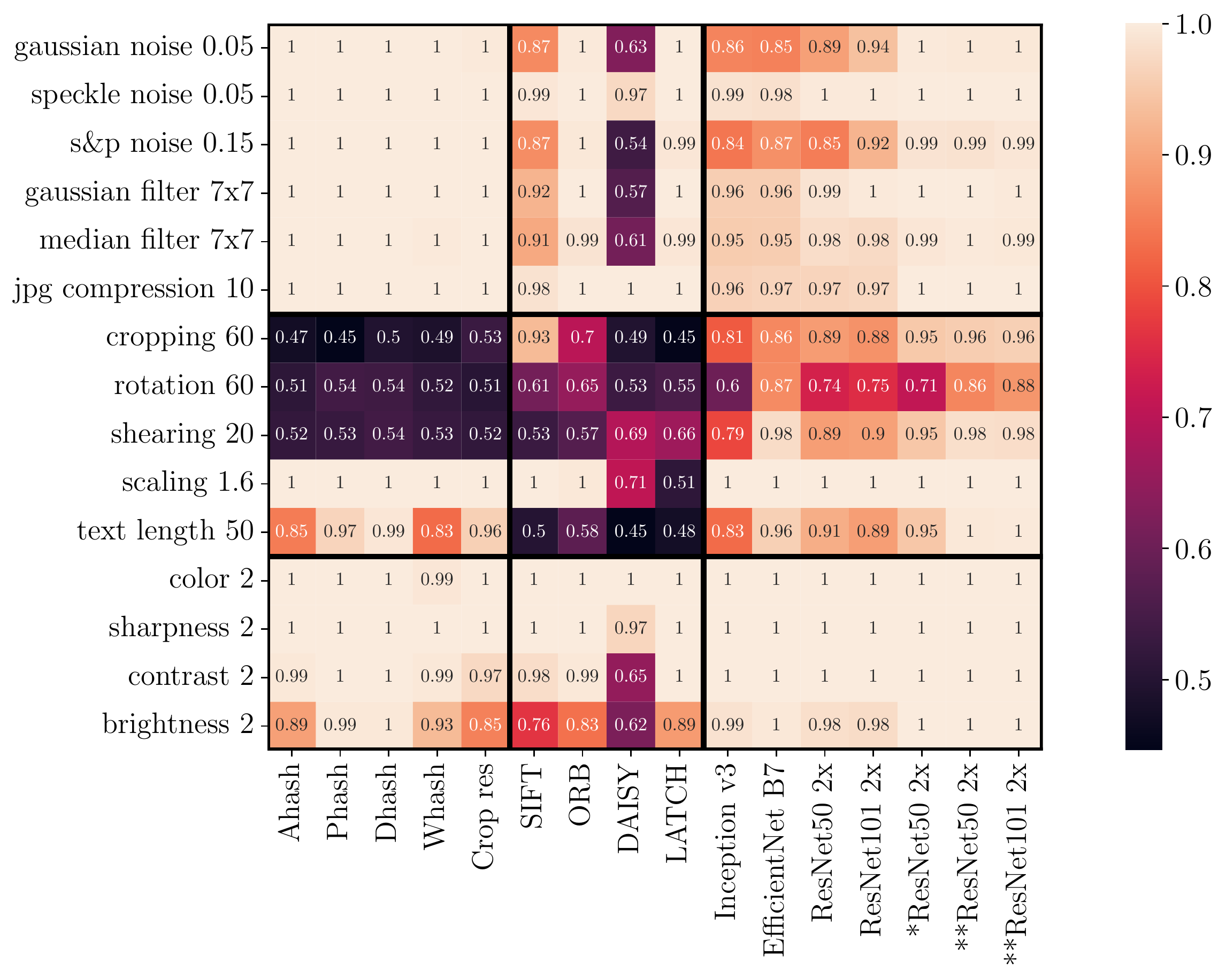}
    \caption{Area under the ROC curve for the maximum intensity of each type of perturbation, for each algorithm.}
    \label{fig:AUC_BSDS500}
\end{figure}

While general performances are interesting, a more detailed analysis is needed to differentiates strength and weaknesses of each method. Figure \ref{fig:AUC_BSDS500} summarizes the AUC for each perturbation type (the strongest intensity of each perturbation). From this, one may observe that noise-like pertubations are very easily dealt with by any block perceptual hashing, while only neural algorithms are able to keep up with strong geometric variations.

\subsection{Influence of database size on false alarms}

\begin{figure} 
    \centering
    \includegraphics[width=\linewidth]{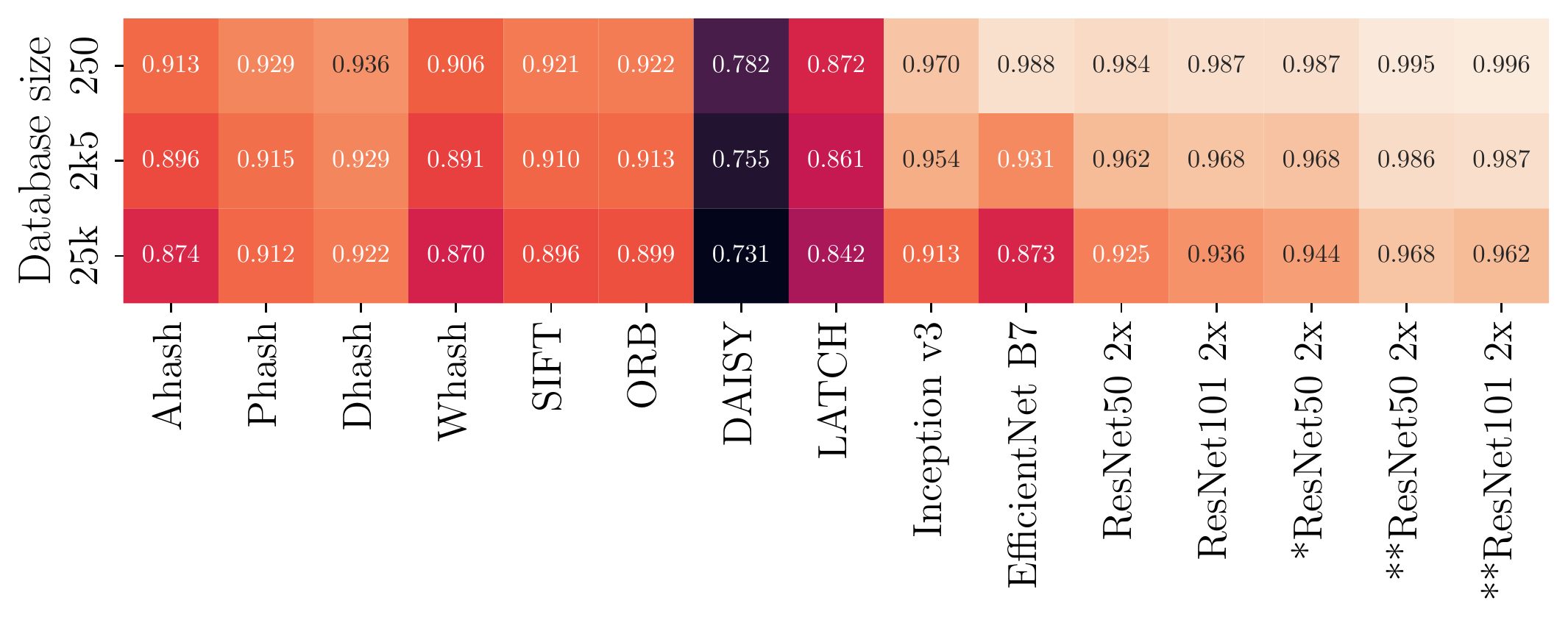}
    \caption{Area under the ROC curve for all perturbations, for each algorithm and different database sizes. The bigger is the database, the worst are the performances}
    \label{fig:AUC_db_size}
\end{figure}

In this part we study the influence of the database size on the performances of the algorithms, with regard to false positive rate.  We use the ImageNet validation set, on which we sample 100 images in each group (control and experimental) and perform the artificial attacks on these. We then use the 100 original images of the experimental group in the database, along with other random images of the experimental group in order to get databases of size 250, 2,500, and 25,000. Since the images to detect are always the same, only the false positive rate is influencing performances on the ROC curves. Figure \ref{fig:AUC_db_size} demonstrates how performances are impacted.

\subsection{Usage on the 2018 in-the-wild Reddit memes dataset}

\begin{figure} 
    \centering
    \subfloat[block perceptual hashing algorithms]{\includegraphics[width=0.5\linewidth]{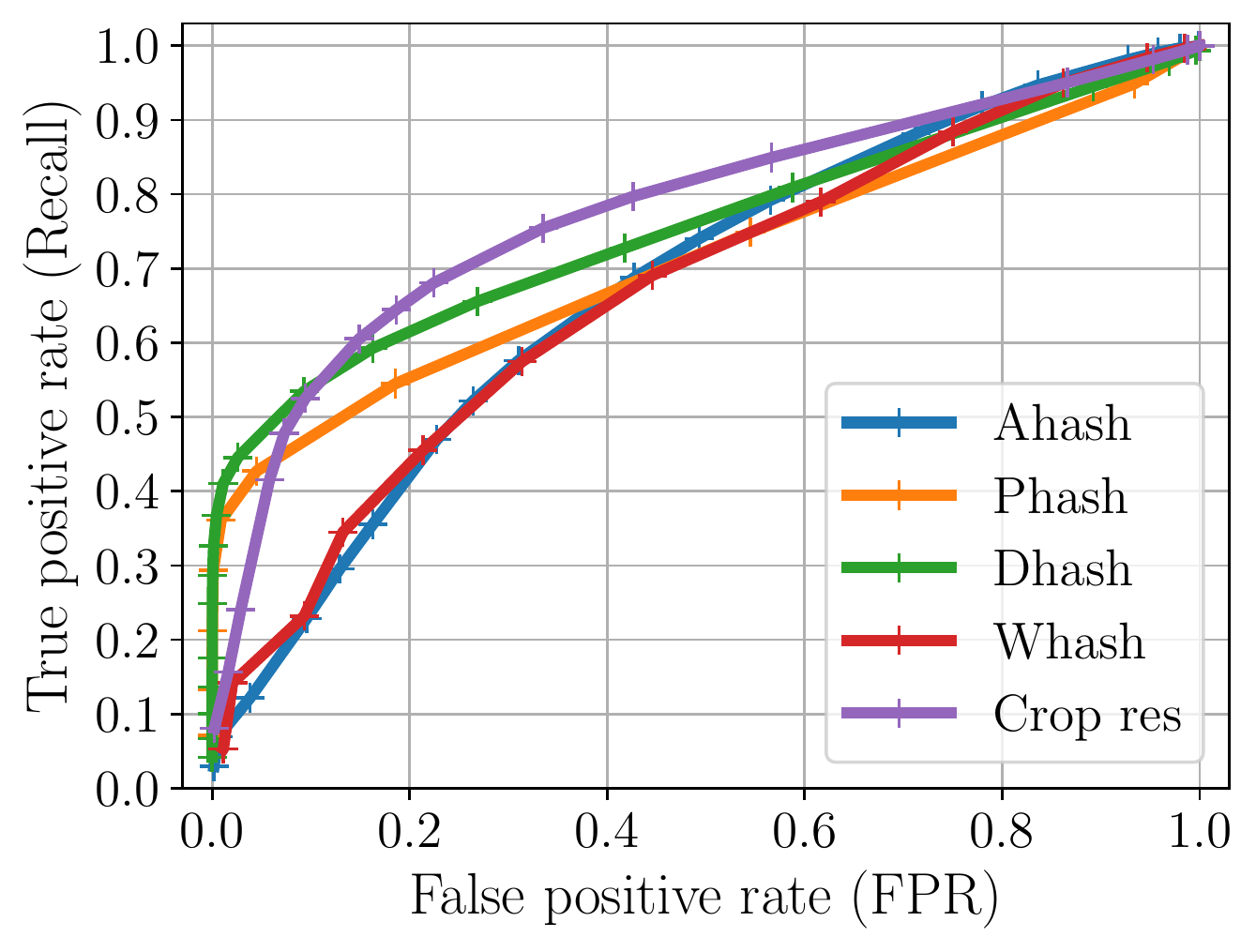}}
    \subfloat[keypoint-based algorithms]{\includegraphics[width=0.5\linewidth]{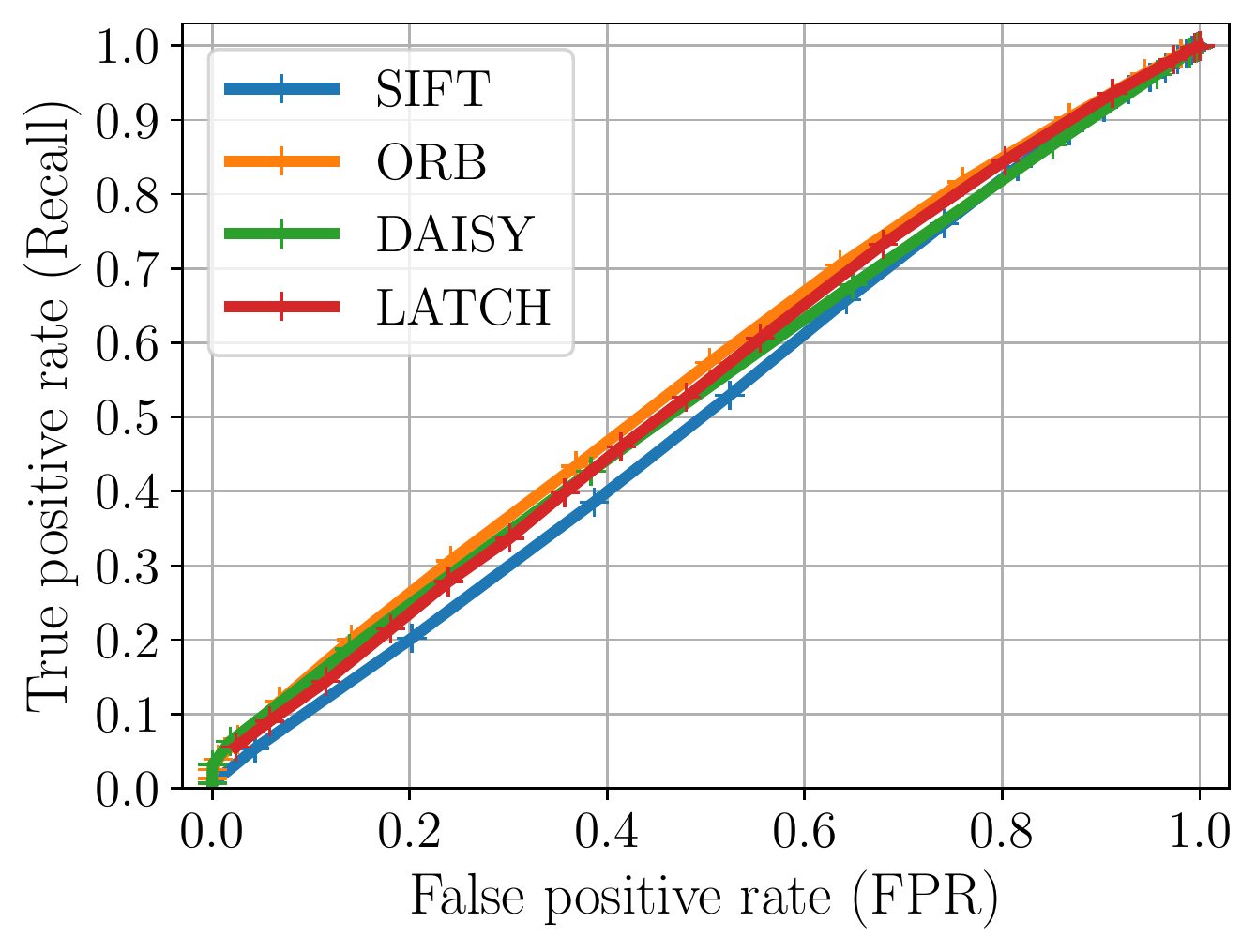}} \\
    \subfloat[neural algorithms trained for classification]{\includegraphics[width=0.5\linewidth]{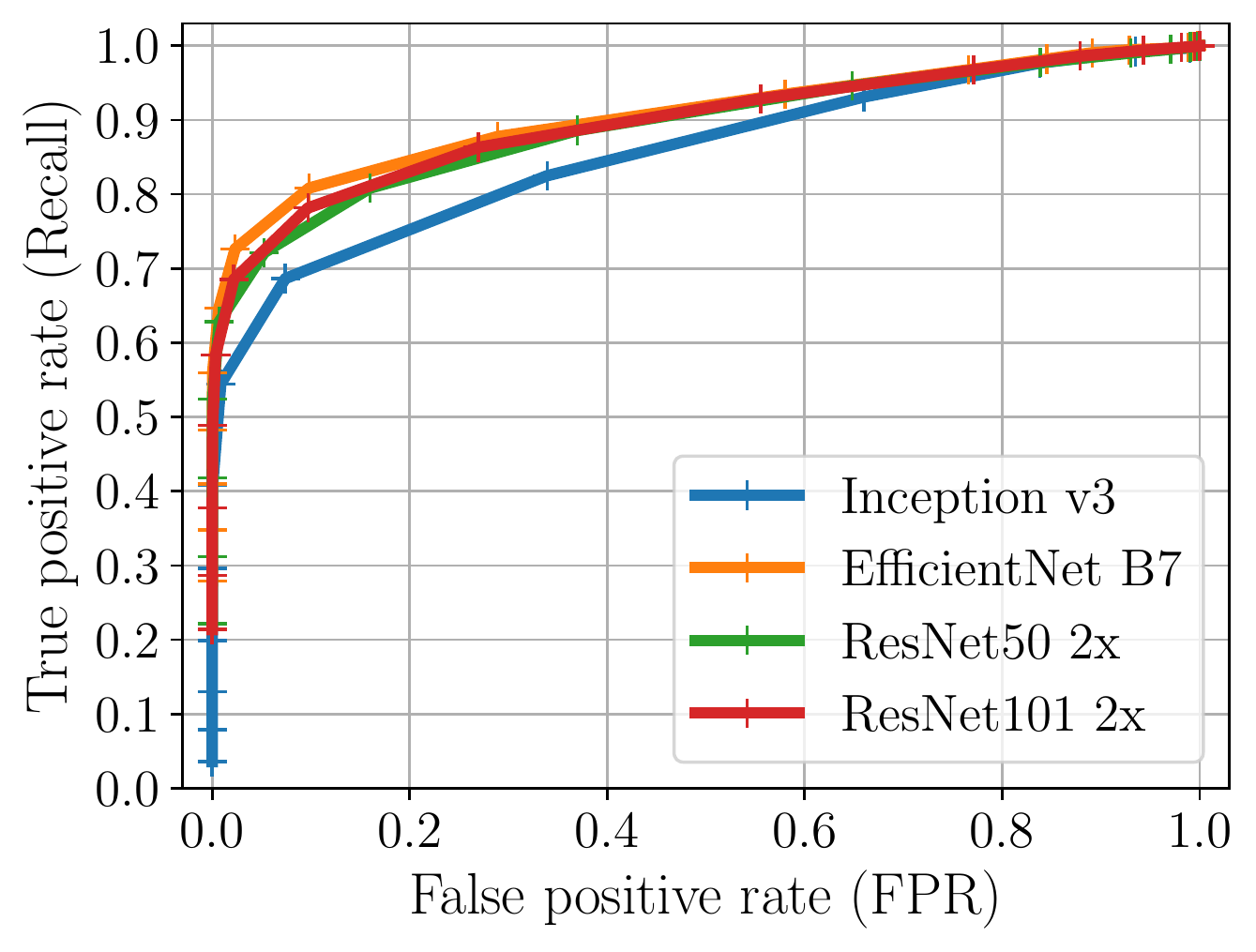}}
    \subfloat[neural algorithms trained in contrastive fashion]{\includegraphics[width=0.5\linewidth]{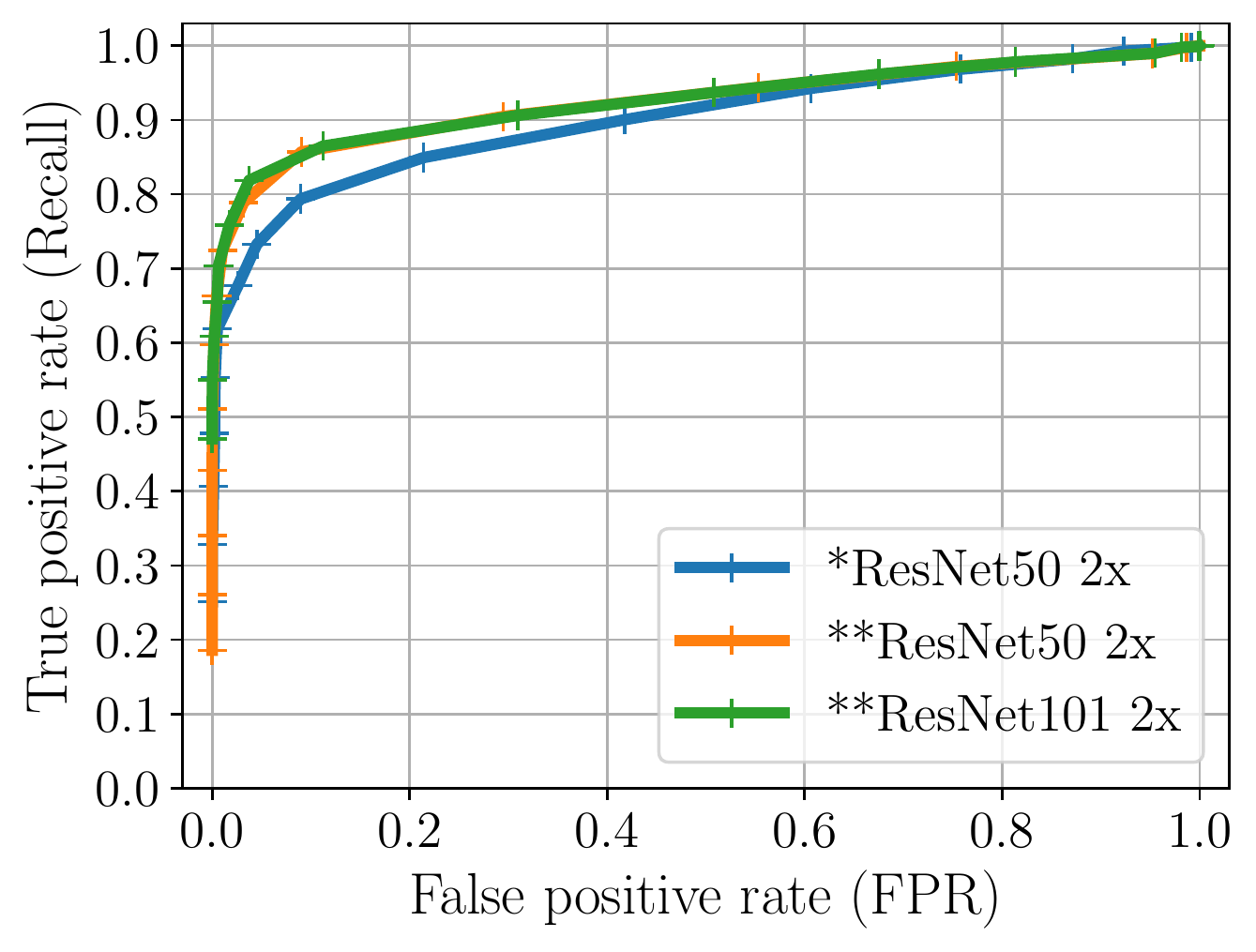}}
    \caption{Overall performances for each algorithms on the Kaggle memes dataset. The parameters of each method are set exactly as for the previous similar study.}
    \label{fig:roc_curves_memes}
\end{figure}

In the following we use the Kaggle dataset of internet "memes". As the database, we use the first occurence of each meme, which is stored as a "meme template" (assuming that most memes semantically differ only by the text, the image staying the same). From this, we hope to detect all variations of this same meme using the above algorithms. Note that this task is harder than artificial perturbations detection as some memes are greatly manipulated by internet users, other than changing text (addition/removal of big objects/persons, swapping large parts of the image with another one, extreme horizontal/vertical stretching of the image,...). Moreover, the template is not always a great representative of the meme, as sometimes it is already a heavily manipulated image compared to the other similar memes. Once again, we split the dataset in two (almost) equal parts. The experimental group is composed of 22,005 images, spanning 128 templates used as the database. The control group contains the remaining 21,655 images, spanning 122 templates. Figure \ref{fig:roc_curves_memes} shows the obtained ROC curves for this dataset. Table \ref{tab:difference_FPR} summarizes the FPR obtained on this memes dataset when using thresholds in distance metric giving about 0.5\% FPR on the BSDS500 dataset. As one is able to see, the FPR does not translate well, and would need to be tuned differently.

\begin{table}[ht]
\centering
\begin{tabular}{|l|c|c|c|}
\hline
Method          & Threshold & FPR on BSDS500 & FPR on Kaggle memes \\ \hline
Ahash           & 0.0520    & 0.72\%         & 6.4\%              \\ \hline
Phash           & 0.2240    & 0.34\%         & 0.90\%              \\ \hline
Dhash           & 0.1590    & 0.65\%         & 0.05\%              \\ \hline
Whash           & 0.0720    & 3.10\%          & 9.30\%             \\ \hline
Crop res        & 0.0690    & 0.36\%         & 5.80\%              \\ \hline
SIFT            & 67.7778   & 0.53\%         & 55.00\%              \\ \hline
ORB             & 0.0906    & 0.55\%         & 21.00\%              \\ \hline
DAISY           & 0.0414    & 0.57\%         & 1.80\%              \\ \hline
LATCH           & 0.1606    & 0.45\%         & 34.00\%              \\ \hline
Inception v3    & 0.2611    & 0.50\%         & 0.00\%                   \\ \hline
EfficientNet B7 & 0.3683    & 0.45\%         & 0.02\%              \\ \hline
ResNet50 2x     & 0.2996    & 0.46\%         & 3.50\%              \\ \hline
ResNet101 2x    & 0.3168    & 0.53\%         & 5.30\%              \\ \hline
*ResNet50 2x    & 0.5197    & 0.55\%         & 13.00\%              \\ \hline
**ResNet50 2x   & 0.5133    & 0.50\%         & 0.79\%              \\ \hline
**ResNet101 2x  & 0.5208    & 0.50\%         & 1.10\%              \\ \hline
\end{tabular}
\vspace{0.3cm}
\caption{False positive rate (FPR) obtained on the Kaggle memes dataset when setting the thresholds to get approximately 0.5\% FPR on the BSDS500 dataset. Note that for the first algorithms, the thresholds are discrete (and thus cannot be perfectly tuned). The corresponding recall can be looked up on receiver-operator curves corresponding to the dataset and method}
\label{tab:difference_FPR}
\end{table}

\section{Discussion}

Unsurprisingly, we observe a wide range of performance between different algorithms calculating perceptual similarity, with certain perturbation types defeating entire classes of algorithms.

Specifically, overall performance of classical computer vision algorithms based on local keypoints proved to be highly susceptible to the addition of meme-like text. This is not surprising - the salient keypoints picked by them are likely to be elements of text rather than elements of background images. Conversely, Dhash algorithm performed surprisingly well across the board, matching or exceeding all other algorithms, except for strong rotation, shearing and cropping - that proved to be difficult to deal with for all algorithms.

Interestingly, Dhash is also the most computationally efficient algorithms, with Phash, performing only slightly worse coming as a close second. SimCLRv2 Resnet50 2x, that performed well across the board  is about 10 times slower than Dhash. This increased performance is not surprising - as we discussed in the introduction regarding Deep Learning, they are able to learn pattern recognition algorithms focusing on salient elements and observe patches of texture as well as their relative position to determine the similartity of the images. It is likely that with additional fine-tuning, SimCLR could perform even better, while remaining just as performant in the inference mode.

Despite the promise of crop resistance, the Crop Resistant Hash was also by far slower than any other method benchmarked, while also being one of the worst performing. It is slower by a factor of about 200x compared to dHash and about 30x to SimCLR Resnet 50 2x. While the idea of the method is attractive, it looks like it doesn't work on the type of media found on the internet to funnel information to public and does not perform in a way that could allow it to deploy at a scale.

Another interesting result of our work concerns the scaling of the length of the block hashes and feature numbers for keypoint algorithms. While block hashing algorithms are virtually insensitive to the hash length in the in performance time, the accuracy gains are not guaranteed. Increasing the hash lenght makes Ahash more accurate, but degrades the accuracy of Phash and has a non-linear effect on the accuracy of dHash. Conversely, increasing the number of keypoints for SIFT and ORB improved their performance as it increased their computational cost almost linearly. At the numbers of keypoints cited by other methods (200-300 keypoints \citep{TechnionCornellReview2022, ProvenancePhilogeny2017PintoScheirer, HybridImageRetrieval2019PhamPark}), they become computationally expensive - almost a 100x worse than dHash and 10x worse than SimCLR. On the other hand, DAISY and LATCH observe absolutely no improvement, likely due to the fact that their edge detection are side-tracked by lines in text that are detected by FAST, as well as non-rotationally robust in rotation and shearing modifications. 

Once again, the relative performance of SIFT is not surprising. It's a keypoint extraction method that focuses on local patches of texture, in a very similar manner as the firs layers of convolutional deep neural networks, minus the relative position matching learnt by the intermediate layers of the latter.

In the similar vein, the relative performance of the Jensen-Shannon distance compared to cosine or L1/L2 distances is a novel result. While widely used in statistical applications and closely related to the cross-entropy, widely used in neural network training, it is rarely used to compare the top-level feature activation for perceptual similarity evaluation. However its intuitive interpretation as the difference of statistics in feature activation makes sense in that context and is coherent with the fact that the Jensen-Shannon distance is one of the top performers across different networks, with no overhead compared to other distances.

Perhaps similarly to the CropHash, the EfficientNet failed to deliver on its promise of lightweight computing. While performing reasonably well, it was about 4-5 times slower than other deep learning methods.

At a scale, we observed a quasiuniversal performance degradation by different methods, with dHash and SimCLR ResNet50 x2 v2 seing the least decay in their performance in the synthetic dataset as the source image bank scaled from 250 to 25 000 images, suggesting they are well adapted to perform at a scale.

This relative performance of Dhash and SimCLR v2 netowrks continued on the in-the-wild memes dataset, where they were among the few ones to observe little to no degradation in performance. Similarly, due to the extensive content remixing found on the in-the-wild memes, the crop-resistant hash pulled ahead, assuming a high (>15\%) FPR was allowed. On real-world data-sets, containing potentially millions of images, this FPR is however unrealistic leading to tens if not hundreds of thousands images being misclassified. Finally, none of classical computer vision algorithms did better than random on the in-the-wild memes dataset.

\section{Conclusion}

In this paper we benchmarked the performance of several common algorithms from three distinct classes for their ability to evaluate a general similarity of two images with modifications likely to occur during their transmission and re-use on the internet.

To our knowledge, our review is the first one to: 
\begin{itemize}
  \item Cover the three classes of algorithms available for overall perceptual image similarity, namely the classic local keypoint and feature ones, Block Hash ones and ones based on Deep Neural Networks
  \item Separately evaluate the robustness of different algorithms to perturbations that would likely be added to images spread on online and social media
  \item Account for the performance difference with a change of hyper parameters
  \item Evaluate the performance of different distance metrics for perceptual vectorization into the real number vectors
  \item Evaluate the computational cost of all the methods presented
\end{itemize}

Our review brought out the following main results:
\begin{itemize}
    \item dHash and SimCLR v2 are best performing across the board, scale well and perform well on real-world datasets. They should be preferred whenever possible. Their performance, computational efficiency and ease of access make their relative absence from research literature surprising and suggest additional scalability and performance are left on the table.
    \item Classical computer vision algorithms based on keypoints and feature extraction do not perform well, do not scale well with the number of features used to match images nor translate to the real-world dataset. It is unsurprising, given that information communication supports on the internet do not conform to the model of deformation they were initially developed for (viewpoint change) and present a large number of modifications they are not robust to, namely text addition.
    \item Jensen-Shannon distance is to be preferred whenever possible for top-level feature activation on deep neural networks. It is just as efficient as common distance metrics and performs better accross the board.
\end{itemize}


We hope that the evaluation of different methods would allow others to judiciously choose the tool that would best suit their needs and computationally capabilities, and would foster the development of new, more efficient tools and generalized benchmarks.

\subsubsection{Acknowledgements} We would like to thank the Armasuisse - Swiss Cyber Defence Campus for the Distinguished Post Doctoral Fellowship supporting AK, and for providing the computational infrastructure for experiments, and Fabien Salvi (EPFL) for the technical support regarding the computational infrastructure organization.

%
%
%
%

\bibliography{anthology,acl2020}
\bibliographystyle{acl_natbib}








\end{document}